%% file: ReviewTemplate.tex
\documentclass[10pt,twocolumn,letterpaper]{article}

\usepackage{cvpr}              

\usepackage{graphicx}
\usepackage{amsmath}
\usepackage{amssymb}
\usepackage{booktabs}

\usepackage{arydshln}
\usepackage{xspace}
\usepackage{adjustbox}

\usepackage{cuted}
\usepackage[subtle]{savetrees}
\usepackage{multirow}

\usepackage[pagebackref,breaklinks,colorlinks]{hyperref}

\usepackage[capitalize]{cleveref}
\crefname{section}{Sec.}{Secs.}
\Crefname{section}{Section}{Sections}
\Crefname{table}{Table}{Tables}
\crefname{table}{Tab.}{Tabs.}

\def\model{MV-GPT\xspace}

\def\cvprPaperID{3222} 
\def\confName{CVPR}
\def\confYear{2022}

\begin{document}

\title{End-to-end Generative Pretraining for Multimodal Video Captioning}

\author{Paul Hongsuck Seo~~~~~~~~Arsha Nagrani~~~~~~~~Anurag Arnab~~~~~~~~Cordelia Schmid\\
Google Research\\
{\tt\small \{phseo,anagrani,aarnab,cordelias\}@google.com}
}

\maketitle

\begin{strip}

    \centering\noindent
    \begingroup
        \captionsetup{type=figure}
        \begin{subfigure}{0.495\textwidth}
            \vspace{-1.9cm}
                        \includegraphics[width=\linewidth]{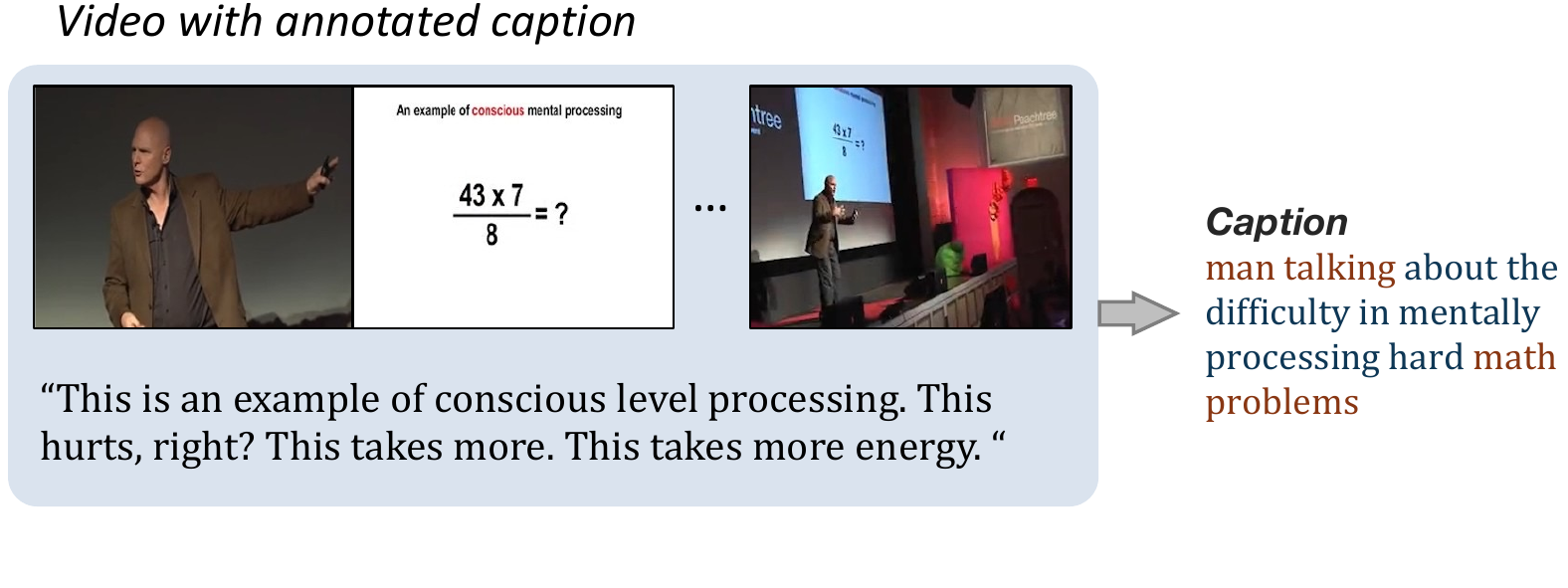}
            \caption{Multimodal video captioning}
            \label{fig:teaser1}
        \end{subfigure}~~~~~
        \begin{subfigure}{0.496\textwidth}
            \vspace{-1.9cm}
                        \includegraphics[width=\linewidth]{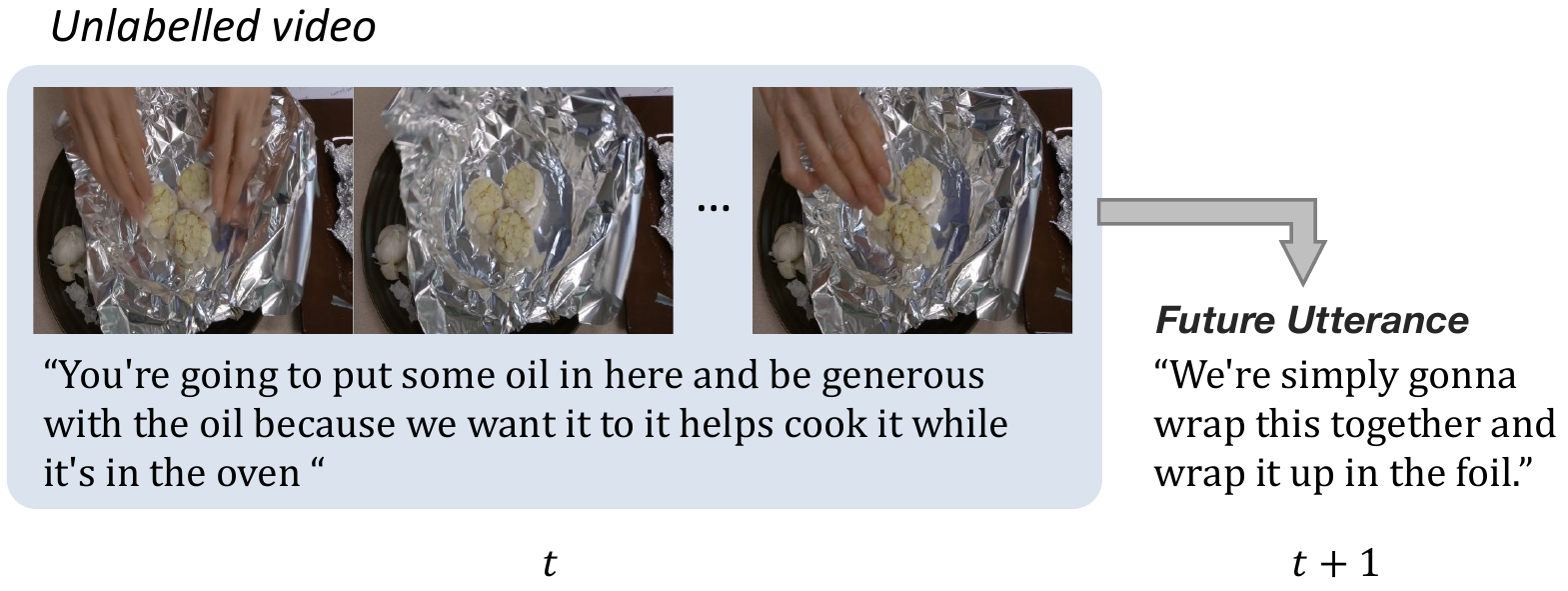}
            \caption{Pretraining using a future utterance}
            \label{fig:teaser2}
        \end{subfigure}
    \endgroup
    \setcounter{figure}{0}
    \vspace{-0.2cm}

    \captionof{figure}{{\bf Generative pretraining for Multimodal Video Captioning.}
    Multimodal Video Captioning takes visual frames and speech transcribed by ASR as inputs and predicts a caption. The example on the left (a) demonstrates that using both modalities jointly is beneficial to generate an accurate caption, \ie, \textcolor[RGB]{137,55,19}{red words} are present in the visual input whereas \textcolor[RGB]{14,55,135}{blue words} correspond to the concepts in the ASR. Our new  multimodal video generative pretraining {\bf (MV-GPT)} uses a {\bf future} utterance in time from the video stream as a captioning target~(b). This objective can be applied to unlabeled data (\eg, HowTo100M), which comes with ASR but no captions, and results in effective joint-pretraining for both a multimodal encoder and decoder.}

    \label{fig:teaser}
    \vspace{-0.2cm}
\end{strip}

\input{sections/abstract}

\input{sections/intro}
\input{sections/related}

\input{sections/method}
\input{sections/experiments}

\input{sections/conclusion}

{\small
\bibliographystyle{ieee_fullname}
\bibliography{egbib}
}

\clearpage
\input{supplementary/appendix}

\end{document}

%% file: sections/abstract.tex
\begin{abstract}
Recent video and language pretraining frameworks lack the ability to generate sentences.
We present Multimodal Video Generative Pretraining (MV-GPT), a new pretraining framework for learning from unlabelled videos which can be effectively used for generative tasks such as multimodal video captioning. 
Unlike recent video-language pretraining frameworks, our framework trains both a multimodal video encoder and a sentence decoder jointly.
To overcome the lack of captions in unlabelled videos, we leverage the future utterance as an additional text source and propose a bidirectional generation objective -- we generate future utterances given the present mulitmodal context, and also the present utterance given future observations.
With this objective, we train an encoder-decoder model end-to-end to generate a caption from raw pixels and transcribed speech directly.
Our model achieves state-of-the-art performance for multimodal video captioning on four standard benchmarks, as well as for other video understanding tasks such as VideoQA, video retrieval and action classification.

\end{abstract}

%% file: sections/intro.tex
\section{Introduction}

A long-standing goal of the AI community is the development of conversational multimodal systems that can both reliably perceive the world and effortlessly communicate with humans. 
An emerging benchmark of progress in this field is the task of multimodal video captioning~\cite{huang2020multimodal,luo2020univl} - which tests both abilities; a successful model must not only accurately understand \textit{`multimodal'} streams of input video (including the speech and the video frames), but also generate coherent natural language descriptions of the content.

Unsurprisingly, a major challenge in the field of vision and language learning is the lack of large-scale, manually annotated data. Annotating captions for videos is time intensive, expensive and subjective (with low inter-annotator agreement~\cite{huang2020multimodal}) -- this is in contrast to fields such as image classification where fully annotated datasets are orders of magnitude larger~\cite{ILSVRC15,hinton2015distilling,yalniz2019billion}. 
To overcome this limitation, there has been a flurry of recent works that pretrain their video-language models on instructional videos~\cite{sun2019videobert,sun2019learning,miech2020end,luo2020univl,seo2021look}, a domain where the speech is particularly well aligned to visual content.
Recently introduced datasets such as Cooking312K~\cite{sun2019videobert} and HowTo100M~\cite{miech2019howto100m} leverage such instructional videos with associated captions from ASR (automatic speech recognition) to learn joint video-and-text embeddings~\cite{sun2019learning,miech2020end} or to train multimodal video encoders~\cite{li2020hero,seo2021look}.
However, the models in these works often do not contain a decoder, lacking the ability to generate sentences, and thus only the video encoder is transferred to the downstream tasks -- indeed for the case of video captioning, the decoder is often learned from scratch~\cite{tang2021decembert,sun2019videobert,zhu2020actbert}.
While one can still initialize the decoder using independently pretrained weights such as those from a GPT-2~\cite{radford2019language} model, we observe that this strategy is suboptimal and performance is significantly improved by optimizing the encoder and the decoder jointly.

\begin{figure*}
    \centering
    \includegraphics[width=\textwidth]{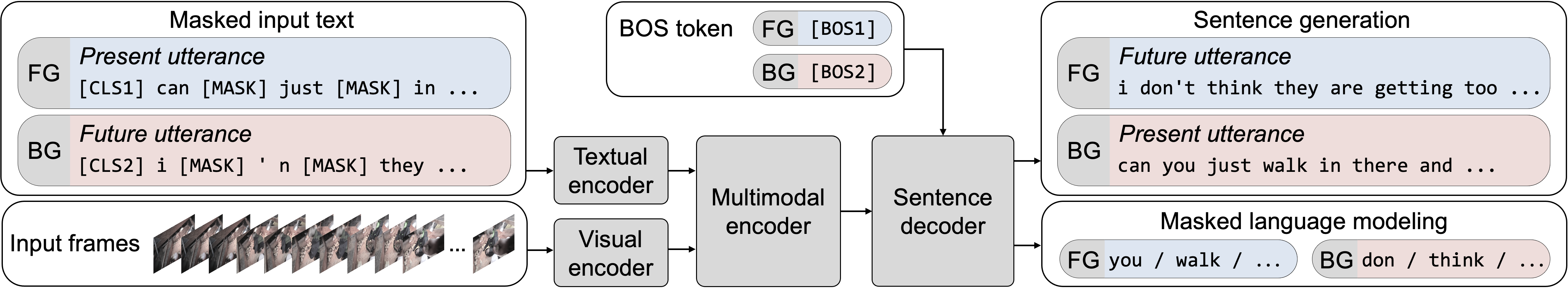}
    \caption{\textbf{Multimodal Video Generative Pretraining (MV-GPT) framework.}
    During pretraining, our network (which consists of 
    modality specific encoders, a multimodal encoder and a sentence decoder)
    is trained with a new bi-directional objective.
    1) Forward generation ({\bf FG}, \colorbox[RGB]{220,230,242}{blue}): Given input frames and present utterances from a video clip, we predict a future utterance and 2) Backward generation ({\bf BG}, \colorbox[RGB]{242,220,219}{red}): Given input frames and a future utterance, predict the current utterances. Both losses are applied to a triplet consisting of video frames, present utterances and a future utterance. 
    To allow our model to recognise the different configurations, we attach distinct special classification tokens \texttt{CLS1} and \texttt{CLS2} to the input text for FG and BG respectively, as well as distinct \texttt{BOS1} and \texttt{BOS2} (beginning of sentence) tokens to the decoder for sentence generation.
	}
    \label{fig:framework}
    \vspace{-0.1cm}
\end{figure*}
For the task of multimodal video captioning, we require a model that can both encode multimodal videos (\ie frames and textual inputs) \textit{and} generate captions. Using multimodal information as input can greatly improve the quality of the generated captions (as illustrated in Figure~\ref{fig:teaser1}). 
However, learning such 
an encoder-decoder model jointly
from unlabelled data is particularly challenging, as it requires two streams of textual data -- naturally occurring transcribed speech accompanying the video for the encoder, and target sentences for the decoder -- whereas unlabelled videos only come with a single stream of speech (Figure~\ref{fig:teaser2}).
Recent works~\cite{korbar2020video,luo2020univl, huang2020multimodal} have attempted to solve this problem 
with a denoising autoencoder - wherein the input speech to the model is artificially \textit{`noised'}, \ie random words are masked out~\cite{korbar2020video,luo2020univl, huang2020multimodal}. The decoder is then tasked with simply reconstructing either the masked phrases or the original unmasked text, where the supervisory signals are provided only from the masked words.
In these frameworks, additional losses are often required to strengthen the pretraining supervision, such as multimodal input alignment~\cite{luo2020univl} and segment ordering~\cite{huang2020multimodal}. 

In our framework, we introduce a novel stronger loss. We leverage future utterances as another source of textual data and train a model to generate these entirely unseen sentences as depicted in Figure~\ref{fig:teaser2}.
To alleviate the problem that future utterances are not temporally aligned,
we propose a
backward generation objective where present aligned utterances are generated given future utterances.
Experimental results show that a model pretrained with this bidirectional generation objective effectively transfers to  multimodal video captioning and outperforms the state of the art by a margin.

We make the following contributions: 
(i) We propose a novel pretraining objective for multimodal video captioning that requires no manually annotated captions, and instead uses utterances sampled at different times in the same video. Our objective is bidirectional in time -- \ie we not only generate future utterances but also the present ones from the future;
(ii) By using two sources of textual data, we are able to jointly train the entire encoder-decoder model. This is unlike previous works which pretrain only the (multimodal) encoder, thereby lacking the ability to generate captions~\cite{seo2021look,sun2019videobert,li2020hero};
(iii) Our encoder is trained from raw pixels and words directly, in contrast with existing methods that rely on pre-extracted visual features limiting transfer to new domains~\cite{luo2020univl,korbar2020video,huang2020multimodal};
(iv) We achieve state-of-the-art results on four video captioning benchmarks -- YouCook2, ViTT, MSR-VTT and ActivityNet-Captions -- 
consistently outperforming existing methods by significant margins; and finally
(v) Our pretraining objective yields strong multimodal video representations, which achieve state-of-the-art performance on other video understanding tasks such as VideoQA, video retrieval and action classification.

%% file: sections/related.tex
\section{Related Work}

\noindent {\bf Video captioning.} Early works in video captioning consisted of rule-based methods~\cite{kojima2002natural,das2013thousand}, where subjects, verbs and objects (SVO-triplets) detected from the video were combined into sentence templates. Later work moved away from rule-based methods by framing captioning as a machine translation task~\cite{rohrbach2013translating,sutskever2014sequence,bahdanau2015neural}, which developed the common encoder-decoder paradigm of today for the task -- the encoder processes a set of video features and accumulates its hidden state, which is then passed to a decoder for
producing a caption.
Early works implemented the visual encoder as a 2D CNN (either frozen or finetuned) applied to video frames, which was then naturally extended to 3D CNNs~\cite{carreira2017quo,xie2018rethinking}, to better capture motion dynamics, with temporal aggregation over the entire video typically performed using attention strategies~\cite{chen2019motion}. Given the computational challenge of using expensive 3D CNNs applied to dense frame inputs (typically 30 fps), most of these works operated on pre-extracted features, only learning the fusion of features in the encoder. Unlike such works, we address this problem using a transformer-based encoder applied to raw pixels~\cite{arnab2021vivit}, sampled at a coarse rate to better capture long range context. 

\noindent {\bf Pretraining with weakly paired data.} 
Existing video captioning datasets~\cite{xu2016msr,zhou2018towards,huang2020multimodal} are orders of magnitude smaller than video classification datasets~\cite{kay2017kinetics}. As a source of weakly paired video and language data, a number of works have used the visual frames and the Automatic Speed Recognition (ASR) transcripts of unlabelled videos to pretrain video representations~\cite{sun2019videobert,sun2019learning,miech2020end,zhu2020actbert,li2020hero,seo2021look}.
These approaches learn multimodal representations by formulating proxy tasks such as masked language/frame modeling~\cite{sun2019videobert,seo2021look}, video-text matching~\cite{miech2020end,li2020hero} or segment ordering~\cite{li2020hero}.
While these studies show improvements on visual representation~\cite{sun2019videobert,sun2019learning,miech2020end,tang2021decembert} or multimodal video representation~\cite{li2020hero,zhu2020actbert,seo2021look} learning, they are designed for discriminative tasks only, and lack the generation capability. 
Pretraining techniques for generative tasks such as ours, are fewer. While \cite{korbar2020video} use multimodal translation as a generative objective, their encoder is limited to accept visual inputs only.
Works that use multimodal inputs to the encoder, train with masking losses -- wherein words or phrases are masked and the objective is to reconstruct the original sentences~\cite{korbar2020video,luo2020univl} or the masked targets~\cite{huang2020multimodal} using an autoregressive generator.
In contrast, we make use of utterances outside of the clip boundary, which are simply ignored in previous works.
We leverage 
future utterances as a second source 
of textual data, and propose a bi-directional generation objective where the model generates the future utterance given the current utterance and vice versa. While we also use a masked language modelling loss, this is simply in addition to our primary generative bidirectional loss.

%% file: sections/method.tex
\section{Method}
Our objective is to pretrain a model that can effectively encode multimodal videos (visual frames and transcribed speech) as well as decode natural language sentences. This will allow us to use the model for multimodal captioning.
In this section, we first describe the pretraining losses used to train the encoder and decoder jointly from unlabelled videos.
We then describe our model, which consists of modality specific encoders, a multimodal encoder and a text decoder (Figure~\ref{fig:framework}).

\subsection{Pretraining Objectives and Losses} 
Our framework is designed to take advantage of unlabelled instructional video data, which consists of video frames and utterances often linked to the visual content~\cite{miech2019howto100m}.
As mentioned earlier, our framework requires two textual streams -- an input to the encoder and a captioning target for the decoder. Because unlabelled videos do not have captioning targets, 
we instead propose a simple objective -- our model is trained to generate a \textit{future} utterance in the video given the current video context and current utterances (forward generation). This gives us two sources of textual supervision, the current utterance allows us to learn how to optimally fuse modalities in the video encoder, while the decoder is tasked with predicting a new utterance it has never seen before. However, our goal is video captioning, and not `predicting the future'. 
To enable our model to generate text corresponding to the present video context, we also add in an additional backward generation loss -- where the model must generate the current utterance given the current video frames and a future utterance (backward generation). This encourages generated sentences to be temporally aligned (and hence more tightly coupled) with the visual inputs.

\subsubsection{Bi-directional Utterance Generation} 
Given a large set of unlabelled videos, we extract short clips consisting of visual frames $F=\{f_1,\dots,f_{N_f}\}$ and transcribed speech utterances $U=\{u_1,\dots,u_{N_u}\}$ aligned with $F$. For each clip, we also consider the immediate future utterance $W=\{w_1,\dots,w_{N_w}\}$ where $u_i$ and $w_j$ are tokenized words in the transcribed utterances. 
Note that we use the term `utterance' to refer to a single sentence of transcribed speech. 
	
\noindent\textbf{Forward Generation:}
Our model is trained to generate a future utterance $W$ given clip frames $F$ and present utterances $U$.
Formally speaking, we formulate our forward generation objective to minimize the negative log-likelihood of the true future utterance $W$, where the loss function given by the chain rule is $\mathcal{L}_\mathrm{FG}=-\sum_{i=1}^{N_w}{\log P(w_i|w_1,\dots,w_{i-1}, F, U)}$.
This loss encourages the pretrained model to effectively encode temporally aligned multimodal inputs to predict the future utterance.

\noindent\textbf{Backward Generation:}
We now apply the same loss as above, albeit in the backward direction. 
Namely, the model is tasked with generating present utterances $U$ aligned with video frames $F$, conditioned on future utterances $W$ and $F$.
As in the forward generation, we also minimize the negative log-likelihood of the true present utterance $\mathcal{L}_\mathrm{BG}=-\sum_{i=1}^{N_u}\log P(u_i|u_1,\dots,u_{i-1}, F, W)$.
Note that the visual input $F$ is temporally aligned with the decoder output $U$.
This loss function encourages the network to generate a caption related to the visual contents.

\subsubsection{Masked Language Modeling}
\label{sec:mlm}
As an additional supplementary loss, we also train with a masked language modeling (MLM) loss~\cite{devlin2018bert}
$\mathcal{L}_\mathrm{MLM}(X)$ where $X$ is the input utterance on which the masking is applied.
We apply this loss on both the forward and backward input utterances, \ie\ as $\mathcal{L}_\mathrm{MLM}(U)$ and $\mathcal{L}_\mathrm{MLM}(W)$.
Note that these losses are computed independently from the above bidirectional generation losses.

Unlike UniVL~\cite{luo2020univl} where the MLM loss is applied to the outputs of the encoder, we apply it to the outputs of the decoder. This encourages the self attention layers in the decoder to focus on further multimodal contextualization of the textual tokens (since each masked token prediction requires knowledge of neighbouring context). As we show in the experiments, this leads to performance gains.

\subsection{Model} \label{sec:architecture}
Our model consists entirely of transformer blocks, and is trained end-to-end directly from pixels and word tokens. 

\subsubsection{Modality Specific Encoders}
Given a multimodal video input consisting of the visual frames $F=\{f_1,\dots,f_{N_f}\}$ and text inputs $X=\{x_1,\dots,x_{N_x}\}$, we first extract features from the individual modalities independently. 
Note here that the textual input $X$ is the aligned utterance $U$ in general (for computing the forward generation loss and for downstream captioning tasks) but is set to $W$ when computing the backward generation loss.

\noindent\textbf{Textual Encoder:} 
We extract $N_x$ contextualized textual embeddings $E=\{e_i\}$ from the input text using a BERT~\cite{devlin2018bert} encoder.

\noindent\textbf{Visual Encoder:} 
Unlike previous approaches~\cite{sun2019videobert, luo2020univl, huang2020multimodal, seo2021look} where visual features are pre-extracted by models pretrained on different datasets, we extract the visual features directly from pixels.
We use the recent transformer-based video encoder ViViT~\cite{arnab2021vivit}, in particular, the tubelet embedding scheme and the factorized encoder architecture. 
For the tubelet embedding scheme we first extract spatio-temporal 3D tubes from the visual input volume resulting in $S\times T$ token embeddings where $S$ and $T$ correspond to the numbers of tokens in the spatial and temporal dimensions, respectively.
Then, the spatial transformer first takes each group of $S$ embeddings from the same temporal index with a special \texttt{CLS} token embedding, and the temporal transformer models interactions between the output \texttt{CLS} embeddings of the individual spatial groups with another \texttt{CLS} embedding resulting in $T+1$ visual features $V=\{v_j\}$ -- see \cite{arnab2021vivit} for further details. 

Unlike 3D CNN visual encoders which operate on consecutive frames extracted at high frame rates (30 fps), our visual encoder can operate on coarsely sampled frames (1 fps), thus significantly reducing compute. This allows us to train the visual encoder end-to-end, and helps adapt our features across the domain gaps between pretraining and downstream datasets. It also allows the easy adoption of off-the-shelf video augmentation directly to RGB frames, which is useful for small-scale downstream benchmarks.

\subsubsection{Multimodal Encoder}
Once the two sets of textual features $E$ and visual features $V$ are extracted, our multimodal encoder fuses multimodal information using the co-attentional transformer used in \cite{lu2019vilbert,seo2021look}.
Each layer consists of two streams where each stream is a stack of two transformer blocks.
In the textual stream, we first contextualize the features $E$ using a cross-attention transformer block attending to the visual features $V$.
Then, the output features are further contextualized by another transformer block with self-attention.
The first transformer block performs inter-modality contextualization through a cross-attention process whereas the second transformer block carries out intra-modality contextualization through a self-attention process.
In the same way, the visual stream $V$ attends to the textual stream. 
Our multimodal encoder repeats this process $R$ times resulting in the output multimodal features $\hat{E}$ and $\hat{V}$.

\subsubsection{Sentence Decoder}
As shown in Figure~\ref{fig:framework}, given multimodal video features $C=\hat{E}\cup \hat{V}$ as context, we autoregressively generate the output sentence $Y$ conditioned on this context using a transformer decoder.
To generate token $y_i$, we first encode the previous generated tokens $Y_i=\{y_{0}, \dots, y_{i-1}\}$ with a look-up table and a positional embedding to produce $H_i=\{h_{0}, \dots, h_{i-1}\}$.
We then encode the context $C$ and the previous embedded tokens $H_i$ using a single transformer. The outputs of this transformer are $\Tilde{C} \cup \Tilde{H}_i$, where $\Tilde{H}_i=\{\Tilde{h}_0,\dots,\Tilde{h}_{i-1}\}$.
Note that $\Tilde{C}$ refers to the multimodal input embeddings obtained from the decoder and is used for computing the MLM loss as discussed in Section~\ref{sec:mlm}.
We then predict the next token $y_i$ from $\Tilde{h}_{i-1}$ by a linear projection with a softmax: $y_i = \mathrm{argmax}(\mathrm{softmax}(\Phi \Tilde{h}_{i-1}))$, where $\Phi \in \mathbb{R}^{\nu\times d}$ is the linear projection matrix and $\nu$ is the vocabulary size.
The first word $h_0$ is set using the special \texttt{BOS} (beginning of sentence) token, and tokens are generated until a special \texttt{EOS} (end of sentence) token is generated.
In practice, each iteration requires only a single forward pass on the decoder transformer with the aid of causal masking introduced in~\cite{vaswani2017attention}. 

\subsubsection{Input and Output Configurations}
\noindent\textbf{Pretraining:} 
Since our pretraining objective is bidirectional, each triplet $(F, U, W)$ consisting of the visual frames $F$, the present utterances $U$ and the future utterance $W$ is processed by the network twice.
For forward generation, the model takes $F$ and $U$ as inputs and generates $W$, and it generates $U$ given $F$ and $W$, in backward generation.
To enable the model to recognize the different configurations, we attach distinct, special tokens \texttt{CLS1} and \texttt{CLS2} to the input text for the forward and backward generation losses respectively as illustrated in Figure~\ref{fig:framework}.
Similarly, we feed distinct \texttt{BOS1} and \texttt{BOS2} tokens to the decoder to initiate sentence generation.

\noindent\textbf{Finetuning for captioning:}
In downstream video captioning datasets, video clips (consisting of frames $F$ and aligned utterances $U$) are manually annotated with a natural language caption.
During finetuning, we attach the \texttt{CLS1} token to $U$ (as is done in forward generation), since $U$ is an aligned utterance, but for generation we feed in the \texttt{BOS2} token (as is done in backward generation to predict the present utterance), so that we also generate a temporally aligned caption.
\vspace{-0.05\baselineskip}

\subsubsection{Implementation Details}

For our text encoder, we adopt the BERT-Base architecture with uncased wordpiece tokenization \cite{devlin2018bert}.
Our visual encoder uses the corresponding ViViT-Base configuration with a 1-layer temporal transformer and a tubelet size of $16\times16\times4$ \cite{arnab2021vivit}.
Our multimodal encoder consists of 2 layers following \cite{seo2021look} and finally, the decoder is based on the GPT-2 (117M parameters) architecture~\cite{radford2019language} but we modify it to take both multimodal input context $C$ and a \texttt{BOS} token allowing conditional generation (the original GPT starts generation immediately by taking the first word as its input and only conditions on text).
We initialize the text encoder and the decoder with the standard BERT and GPT-2 weights respectively pretrained on large-scale unlabelled corpora \cite{devlin2018bert,radford2019language}.
Similarly, we initialize our visual encoder using the pretrained weights on Kinetics 400 in \cite{arnab2021vivit} unless otherwise specified. Our model is pretrained end-to-end using the Adam optimizer~\cite{kingma2014adam} for 1.5M iterations with the batch size of 2048.
For more detailed hyperparameters and training strategies for pretraining and finetuning, please refer to the appendix.

%% file: sections/experiments.tex
\section{Experiments}
In this section, we first demonstrate our results on four different benchmarks for multimodal video captioning. We then also show that our pretrained model has the ability to generalise to other video understanding tasks such as video question answering (VideoQA), video retrieval and action classification.

\subsection{Multimodal Video Captioning}
\label{sec:exp_captioning}

\subsubsection{Datasets and Evaluation Protocols}
We use HowTo100M~\cite{miech2019howto100m} as our pretraining dataset, and evaluate on four downstream captioning benchmarks. \\ 
\noindent\textbf{HowTo100M}~\cite{miech2019howto100m} consists of 1.2M instructional videos from YouTube. Transcribed speech is obtained using the YouTube ASR API~\cite{youtubeapi}.
Following \cite{seo2021look}, we extract 53M triplets of frames, current utterances and future utterances for pretraining.

\noindent\textbf{YouCook2}~\cite{zhou2018towards} is the most widely adopted benchmark for multimodal video captioning and contains 2,000 cooking videos for 89 different dishes with 14K video clips. 
Each video clip is annotated with a single captioning sentence.

\noindent\textbf{Video Timeline Tags (ViTT)}~\cite{huang2020multimodal} 
was created to better reflect the distribution of instructional videos in the wild. 
It consists of 8,169 videos, 5,840 of these videos for training and the remaining videos for validation and testing. 
Videos are divided into 7.1 segments on average, with each segment accompanied by a short timeline tag.

\noindent\textbf{MSR-VTT}~\cite{xu2016msr} is a standard benchmark with 10K open domain video clips for video captioning.
The duration of each video clip is between 10 and 30 seconds, and 20 natural language descriptions are manually annotated per clip.

\noindent\textbf{ActivityNet-Captions}~\cite{krishna2017dense} is a standard dense video captioning benchmark consisting of 100K temporally localized
sentences for 20k videos. 
We follow the standard splits with 50/25/25\% examples for training, validation and test sets.
To evaluate our model's ability to predict captions, we use ground truth temporal proposals following~\cite{krishna2017dense}.

We pretrain a single model on HowTo100M, which is then transferred to all four captioning benchmarks through finetuning. 
We report results using the following established metrics: BLEU-4 (B-4)~\cite{papineni2002bleu}, CIDEr (C)~\cite{vedantam2015cider}, METEOR (M)~\cite{banerjee2005meteor} and ROUGE-L (R-L)~\cite{lin2004rouge}. 
For ViTT, we measure BLEU-1 (B-1) instead of BLEU-4 following \cite{huang2020multimodal}.

\subsubsection{Results}
In this section we ablate some key design choices, in particular the backbone and objective functions used in MV-GPT, and explore the impact of the end-to-end training.
Finally, we compare our model to the state of the art.

\begin{table}[t]
    \centering
    \scalebox{0.85}{
     \begin{tabular}{@{\hskip 0.8mm}l@{\hskip 3.5mm}c@{\hskip 3.5mm}c@{\hskip 3.5mm}c@{\hskip 3.5mm}c@{\hskip 3.5mm}c@{\hskip 0.8mm}}
    \toprule
        PT Losses  & PT parts & B-4 & C & M & R-L \\
    \midrule 
 No PT & -- & 13.25 & 1.03 & 17.56 & 35.48 \\
 Baseline PT & E & 16.13 & 1.46 & 21.76 & 41.50 \\
 CoMVT~\cite{seo2021look} & E & 14.46 & 1.24 & 18.46 & 37.17 \\
 M-MASS~\cite{huang2020multimodal} & E+D & 19.03 & 1.88 & 24.00 & 45.10 \\
 UniVL~\cite{luo2020univl} & E+D & 19.95 & 1.98 & 25.27 & 46.81 \\
 MV-GPT (Ours) & E+D & \bf21.26 & \bf2.14 & \bf26.36 & \bf48.58 \\ 
    \bottomrule
    \end{tabular}}
    \caption{
    Comparisons to existing pretraining losses on YouCook2. 
    \textbf{PT} stands for pretraining. \textbf{PT parts} indicates which part of the model are pretrained, encoder (E) or both encoder and decoder (E + D).
    We reimplement the loss functions of existing methods but use our model and training strategies for fair comparison. 
    } 
    \label{tab:comps-loss}
\end{table}

\noindent\textbf{Pretraining Losses:} 
We implement a simple baseline, which consists of a masked language modelling loss given visual frames and ASR as input (Baseline PT). 
We also reimplement three state-of-the-art pretraining losses: (i)~CoMVT~\cite{seo2021look}, (ii)~UniVL~\cite{luo2020univl} and (iii)~M-MASS~\cite{huang2020multimodal}. For a fair comparison, we use our model architecture for all experiments, varying the loss function only. For the methods which pretrain the encoder only, we initialise the decoder with public GPT-2 weights~\cite{radford2019language}. 
For `No PT', the encoder is not pretrained either, but is initialized with public BERT and ViViT pretrained on ImageNet21k.

Table~\ref{tab:comps-loss} compares these different losses. We can observe that  pretraining the encoder only brings moderate gains over training from scratch, for all the losses investigated.
This performance is greatly improved by pretraining both the encoder and decoder jointly. Finally, we observe that our approach MV-GPT outperforms existing joint pretraining losses.

\begin{table}[t]
    \centering
    \scalebox{0.85}{
     \begin{tabular}{@{\hskip 0.8mm}c@{\hskip 3mm}c@{\hskip 3mm}c@{\hskip 3mm}c@{\hskip 3mm}c@{\hskip 3mm}c@{\hskip 3.5mm}c@{\hskip 3.5mm}c@{\hskip 3.5mm}c@{\hskip 0.8mm}}
    \toprule
        FG & BG & MLM-E & MLM-D & WD & B-4 & C & M & R-L \\
    \midrule 
 \multicolumn{5}{c}{\textit{No PT}}& 13.25 & 1.03 & 17.56 & 35.48 \\
  & & \checkmark & & & 16.13 & 1.46 & 21.76 & 41.50 \\ 
 \checkmark & & \checkmark & & & 20.65 & 2.05 & 25.81 & 47.22 \\ 
 \checkmark & & & \checkmark & & 20.77 & 2.09 & 25.90 & 47.41 \\
 \checkmark & \checkmark & \checkmark & & & 20.82 & 2.10 & 26.20 & 48.22 \\ 
 \checkmark & \checkmark & & \checkmark & &  20.89 & 2.11 & 26.42 & 48.30 \\ 
 \checkmark & \checkmark & & \checkmark & \checkmark & \bf21.26 & \bf2.14 & \bf26.36 & \bf48.58 \\ 
    \bottomrule
    \end{tabular}}
    \caption{Ablation on YouCook2 showing the effect of our different loss components in pretraining. \textbf{FG:} Forward Generation loss. \textbf{BG:} Backward Generation loss. \textbf{MLM-E/MLM-D:} Masked Language Modelling loss applied on encoder outputs (E) or decoder outputs (D). \textbf{WD:} Weight Decay. \textbf{No PT:} No pretraining with any of these losses.
    } 
    \label{tab:abl-losses}
\end{table}
\noindent\textbf{Effect of each Loss Term in MV-GPT:}
Table~\ref{tab:abl-losses} shows the effect of each term in our loss function. 
The forward generation (FG) loss already provides strong supervision.
When applying the masked language modelling loss on the decoder outputs (MLM-D) instead of the encoder outputs (MLM-E), performance is slightly improved due to the additional input contextualization provided by the decoder.
Adding the backward generation (BG) loss provides a boost across all metrics.  
Additionally, we observe that adding weight decay (WD)~\cite{krogh1992simple} brings additional gains, and we report our scores in this full setting for the rest of the paper.

\begin{table}[t]
    \centering
    \scalebox{0.76}{
    \begin{tabular}{@{\hskip 0.8mm}c@{\hskip 1.5mm}c@{\hskip 3mm}c@{\hskip 2mm}c@{\hskip 3mm}c@{\hskip 3mm}c@{\hskip 3mm}c@{\hskip 3mm}c@{\hskip 0.8mm}}
    \toprule
        \multirow{2}{*}{Arch.}  & \multirow{2}{*}{Weights from / Trained on} & \multicolumn{2}{@{\hskip 0mm}c@{\hskip 3mm}}{E2E} & \multirow{2}{*}{B-4} & \multirow{2}{*}{C} & \multirow{2}{*}{M} & \multirow{2}{*}{R-L} \\
        &&PT&FT\\
    \midrule 
    \multicolumn{8}{c}{\textit{YouCook2}} \\
    \midrule
S3D & S3D \cite{xie2018rethinking} / Kinetics & & & 19.65 & 1.93 & 24.47 & 45.79 \\
S3D & MIL-NCE \cite{miech2020end} / HowTo100M & & & 20.02 & 1.96 & 24.98 & 46.65 \\ 
 \hdashline
ViViT & ViViT \cite{arnab2021vivit} / Kinetics & & & 19.54 & 1.93 & 24.42 & 45.93 \\
ViViT & MV-GPT / HowTo100M & \checkmark & & 21.77 & 2.20 & 26.97 & 49.29 \\
ViViT & MV-GPT / HowTo100M & \checkmark & \checkmark & 21.26 & 2.14 & 26.36 & 48.58 \\
ViViT & MV-GPT / HowTo100M & \checkmark & \checkmark$\dagger$ & \bf21.88 & \bf2.21 & \bf27.09 & \bf49.38 \\
    \midrule
    \multicolumn{8}{c}{\textit{MSR-VTT}} \\
    \midrule
ViViT & MV-GPT / HowTo100M & \checkmark & & 47.04 & 0.55 & 36.80 & 62.99 \\
ViViT & MV-GPT / HowTo100M & \checkmark & \checkmark & \bf48.92 & \bf0.60 & \bf38.66 & \bf64.00 \\
    \bottomrule
    \end{tabular}}
    \caption{Ablation on YouCook2 with different visual encoder configurations.
    \textbf{E2E:} End-to-end training including the visual encoder.
    \textbf{PT:} Pretraining.
    \textbf{FT:} Finetuning.
    $\dagger$~Freeze the visual encoder at the beginning and tune end-to-end once converged during finetuning.
    }
    \label{tab:abl-backbone}
\vspace{-0.3cm}
\end{table}

\noindent\textbf{Visual Encoder and End-to-end Training:}
In Table~\ref{tab:abl-backbone}, we first compare the ViViT~\cite{arnab2021vivit} encoder to commonly used S3D features~\cite{xie2018rethinking}.
When both encoders are trained on Kinetics and fixed for multimodal pretraining and finetuning, they show comparable scores despite the large complexity of S3D due to the high frame rate required (30 fps vs.\ 1 fps for ViViT).
Using HowTo100M to train a visual encoder, we observe large gains with both architectures as expected given the similarity in the domains -- HowTo100M and YouCook2 are both instructional video datasets.
However, we observe larger gains with ViViT where the visual encoder is optimized for  generative losses within our framework and jointly trained with the other components thanks to the low complexity of the ViViT encoder. 
These results show the benefits of end-to-end pretraining.

We further investigate the effects of end-to-end training for finetuning. 
For YouCook2, we observe slight performance degradation when naively finetuning the network end-to-end from the beginning (row 4 to 5).
This degradation is overcome by initially freezing the visual encoder and starting end-to-end training after convergence, which gives us a minor gain (row 6).
These results indicate that our pretrained visual encoder already captures strong representations for inputs in a similar domain, and end-to-end finetuning is less critical in this case.
However, we observe more significant gains on MSR-VTT since end-to-end finetuning becomes crucial given a larger domain gap (row 7 to 8).

\begin{table}[t]

\vspace{-0.1cm}
    \centering
    \scalebox{0.85}{
    \begin{tabular}{@{\hskip 0.8mm}cc@{\hskip 2mm}c@{\hskip 2mm}c@{\hskip 2mm}c@{\hskip 2mm}c@{\hskip 0.8mm}}
    \toprule
    Initialization & MV-GPT Pretraining & B-4 & C & M & R-L \\ 
    \midrule 
        Random & & 10.93 & 64.56 & 12.88 & 29.03  \\
        Random &$\checkmark$ & 20.78 & 2.09 & 25.83 & 47.76   \\
        Public weights & & 13.25 & 1.03 & 17.56 & 35.48\\
        Public weights &$\checkmark$ & \bf21.26 & \bf2.14 & \bf26.36 & \bf48.58   \\
    \bottomrule
    \end{tabular}}
    \caption{Ablations on YouCook2 showing the effect of initialization and pretraining. \textbf{Public Weights}: Initialization with public BERT, GPT-2 and ViViT weights.
    } 
    \label{tab:scratch}
\end{table}

\noindent\textbf{Pretraining with Random Initialization:}
We also investigate the ability of the model to learn from scratch.
We initialize the model either entirely randomly or using pretrained BERT, ViViT and GPT-2 weights.
Table~\ref{tab:scratch} shows that with random initialization, our method still performs very well (row 2), outperforming the model initalized with public BERT, GPT-2 and ViViT weights (row 3). Note that the pretrained ViViT weights were obtained from training on the fully supervised dataset Kinetics. 
Also, pretraining entirely from scratch even approaches the case where all parts of the model are intialized using public weights and pretrained (row 4).

\noindent\textbf{Multimodal vs.\ Single Modality:}
In Table~\ref{tab:sota-youcook}, we show results with text only and visual only inputs (we only feed the \texttt{CLS} token for the omitted modality). It is clear that both modalities are complementary and performance is best when combining both.
Additionally, to assess the contribution of the visual modality, we test a model pretrained with text inputs only. Even when this pretrained model is finetuned with both modalities, the performance is significantly lower compared to a pretrained multimodal model (last row in Table~2): there is a 25\% relative drop on all 4 metrics (\eg, 1.43 vs. 2.14 in CIDEr).
When finetuned with text inputs only, the scores drop further (\eg, to 1.20 in CIDEr). These results confirm the importance of the visual inputs during pretraining.

\begin{table}[t]
    \centering
    \scalebox{0.85}{
    \begin{tabular}{@{\hskip 0.8mm}l@{\hskip 1mm}c@{\hskip 2.5mm}c@{\hskip 3mm}c@{\hskip 3mm}c@{\hskip 3mm}c@{\hskip 3mm}c@{\hskip 0.8mm}}
    \toprule
        Method & PT parts & Inputs & B-4 & C & M & R-L \\
    \midrule 
        VideoBERT~\cite{sun2019videobert} & E & V & 4.04 & 0.49 & 11.01 & 27.50 \\
        ActBERT~\cite{zhu2020actbert} & E & V & 5.41 & 0.65 & 13.30 & 30.56 \\
        MART~\cite{lei2020mart} & -- & V & 8.00 & 0.36 & 15.90 & --\\
        AT~\cite{hessel2019case} & -- & T & 8.55 & 1.06 & 16.93 & 35.54 \\ 
        \hdashline
        DPC~\cite{shi2019dense} & -- & V+T & 2.76 & -- & 18.08 & -- \\
        AT+Video~\cite{hessel2019case} & -- & V+T & 9.01 & 1.12 & 17.77 & 36.65 \\ 
        DECEMBERT~\cite{tang2021decembert} & E & V+T & 11.92 & 0.58 & 20.01 & 40.22 \\
        VideoAsMT~\cite{korbar2020video} & E+D & V & 5.30 & -- & 13.40 & --\\
        M-MASS~\cite{huang2020multimodal} & E+D & V+T & 12.04 & 1.23 & 18.32 & 39.03 \\
        UniVL~\cite{luo2020univl} & E+D & V+T & 17.35 & 1.81 & 22.35 & 46.52 \\ 
        \hdashline
        MV-GPT (Ours) & E+D & V & 16.71 & 1.53 & 21.43 & 41.56 \\ 
        MV-GPT (Ours) & E+D & T & 16.71 & 1.56 & 20.88 & 40.19 \\ 
        \bf MV-GPT (Ours) & \bf E+D & \bf V+T & \bf 21.88 & \bf2.21 & \bf 27.09 & \bf 49.38 \\
    \bottomrule
    \end{tabular}}
    \caption{Comparison to SOTA on YouCook2 for video captioning. 
    }
    \label{tab:sota-youcook}
\end{table}

\begin{table}[t]
    \centering
    \scalebox{0.85}{
    \begin{tabular}{@{\hskip 0.8mm}l@{\hskip 1mm}c@{\hskip 2.5mm}c@{\hskip 3mm}c@{\hskip 3mm}c@{\hskip 3mm}c@{\hskip 3mm}c@{\hskip 0.8mm}}
        \toprule
        Method & PT parts & Inputs&  B-1 & C & M & R-L \\
        \midrule
        M-MASS~\cite{huang2020multimodal} & E+D & V+T & 22.37 & 0.82 & 11.00 & 31.40\\ 
        \bf MV-GPT (Ours) & \bf E+D & \bf V+T & \bf 37.89 & \bf 1.04 & \bf 26.75 & \bf 34.76 \\ 
        \bottomrule
    \end{tabular}}
    \caption{Comparison to SOTA on ViTT for video captioning.}
    \label{tab:sota-vitt}
\end{table}

\begin{table}[t]
    \centering
    \scalebox{0.85}{
    \begin{tabular}{@{\hskip 0.8mm}l@{\hskip 1mm}c@{\hskip 2.5mm}c@{\hskip 3mm}c@{\hskip 3mm}c@{\hskip 3mm}c@{\hskip 3mm}c@{\hskip 0.8mm}}
    \toprule
        Method & PT parts & Inputs & B-4 & C & M & R-L \\
        \midrule 
        OA-BTG~\cite{zhang2019object} & -- & V & 41.40 & 0.47 & 28.20 & -- \\
        MGSA~\cite{chen2019motion} & -- & V & 42.40 & 0.48 & 27.60 &  -- \\
        POS+CG~\cite{wang2019controllable} & -- & V & 42.00 & 0.49 & 28.20 & 61.60 \\
        POS+VCT~\cite{hou2019joint} & -- & V & 42.30 & 0.49 & 29.70 & 62.80 \\
        SAM-SS~\cite{chen2020semantics} & -- & V & 43.80 & 0.51 & 28.90 & 62.40 \\
        ORG-TRL~\cite{zhang2020object} & -- & V & 43.60 & 0.51 & 28.80 & 62.80 \\
        VNS-GRU~\cite{chen2020delving} & -- & V & 45.30 & 0.53 & 29.90 & 63.40\\ \hdashline
        DECEMBERT~\cite{tang2021decembert} & E & V & 45.20 & 0.52 & 29.70 & \bf 64.70 \\ 
        VideoAsMT~\cite{korbar2020video} & E+D & V & 41.70 & -- & 28.50 & --\\
        UniVL~\cite{luo2020univl} & E+D & V+T & 41.79 & 0.50 & 28.94 & 60.78 \\
        \hdashline
        \bf MV-GPT (Ours) & \bf E+D & \bf V+T & \bf 48.92 & \bf 0.60 & \bf 38.66 & \bf64.00 \\ 
        \bottomrule
    \end{tabular}}
    \caption{Comparison to SOTA on MSR-VTT for video captioning. }
    \label{tab:sota-msr}
\end{table}

\begin{table}[t]
    \centering
    \scalebox{0.85}{
    \begin{tabular}{@{\hskip 0.8mm}lcccc@{\hskip 0.8mm}}
    \toprule
        Method & B-4 & M \\
        \midrule 
        DCEV~\cite{krishna2017dense} & 1.60 & 8.88 \\
        DVC~\cite{li2018jointly} & 1.71 & 9.31 \\
        Bi-SST~\cite{wang2018bidirectional}& -- & 10.89 \\
        HACA~\cite{wang2018watch} & 2.71 & 11.16 \\
        MWSDEC~\cite{rahman2019watch} & 1.46 & 7.23 \\
        MDVC~\cite{iashin2020multi} & 1.46 & 7.23 \\
        BMT~\cite{iashin2020better} & 1.99 & 10.90 \\\hdashline
        \bf MV-GPT (Ours) & \bf 6.84 & \bf 12.31 \\ 
        \bottomrule
    \end{tabular}}
    \caption{
    Comparison to SOTA on ActivityNet-Captions for video captioning with ground-truth action proposals.}
    \label{tab:sota-activitynet}
\end{table}

\begin{figure*}
    \centering
    \scalebox{0.85}{
        \begin{tabular}{@{\hskip 0mm}c@{\hskip 4mm}c@{\hskip 0.8mm}}

            \begin{tabular}{@{\hskip 0mm}p{8.5cm}@{\hskip 2mm}p{3.3cm}@{\hskip 0.8mm}}
                \centering
                \adjincludegraphics[valign=c,width=0.48\linewidth]{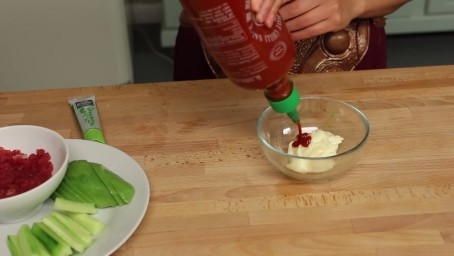}~
                \adjincludegraphics[valign=c,width=0.48\linewidth]{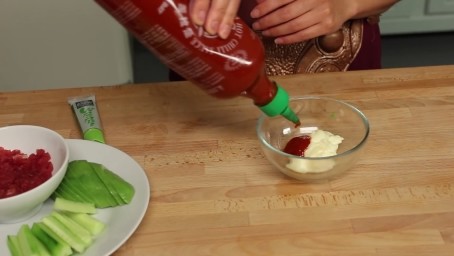}
                &\parbox[c]{1\linewidth}{\textbf{Transcript}: 
                
                This makes a really good source. So about twenty five spice you like it. That's about 4 teaspoons}
                
            \end{tabular}
            & 
            \bgroup
            
            \begin{tabular}{@{\hskip 0mm}p{8.2cm}@{\hskip 0.8mm}}
                \raggedright \textbf{Generated captions}  \tabularnewline[1.2mm]
                \raggedright ~\colorbox[RGB]{220,230,242}{GT:} pour in spicy sauce \tabularnewline[1mm]
                \raggedright ~\colorbox[RGB]{242,220,219}{No-PT:} pour some sauce over the pasta \tabularnewline[1mm]
                \raggedright ~\colorbox[RGB]{235,241,222}{MV-GPT:} add sriracha to the bowl \tabularnewline
            \end{tabular}
            \egroup \tabularnewline[1.2cm]
            
            \begin{tabular}{@{\hskip 0mm}p{8.5cm}@{\hskip 2mm}p{3.3cm}@{\hskip 0.8mm}}
                \centering
                \adjincludegraphics[trim={0 1.5cm 0 0},clip,valign=c,width=0.48\linewidth]{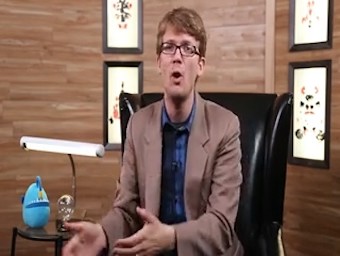}~
                \adjincludegraphics[trim={0 1.5cm 0 0},clip,valign=c,width=0.48\linewidth]{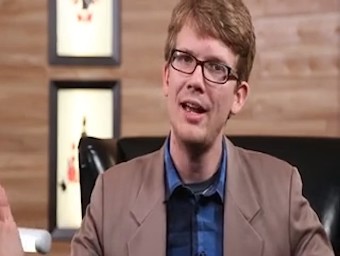}
                &\parbox[c]{1\linewidth}{\textbf{Transcript}: 
                
                So by considering the whole host of nature and nurture influences, we can take a broader view of mental health ...}
                
            \end{tabular}
            & 
            \bgroup
            
            \begin{tabular}{@{\hskip 0mm}p{8.2cm}@{\hskip 0.8mm}}
                \raggedright \textbf{Generated captions}  \tabularnewline[1.2mm]
                \raggedright ~\colorbox[RGB]{220,230,242}{GT:} a man in a brown blazer discussing 
                %
                mental health \tabularnewline[1mm]
                \raggedright ~\colorbox[RGB]{242,220,219}{No-PT:} a man in a blue shirt is talking \tabularnewline[1mm]
                \raggedright ~\colorbox[RGB]{235,241,222}{MV-GPT:} a man in a suit is talking about mental health \tabularnewline
            \end{tabular}
            \egroup  \tabularnewline[1.25cm]

            \begin{tabular}{@{\hskip 0mm}p{8.5cm}@{\hskip 2mm}p{3.3cm}@{\hskip 0.8mm}}
                \centering
                \adjincludegraphics[trim={0 1.5cm 0 0.5cm},clip,valign=c,width=0.48\linewidth]{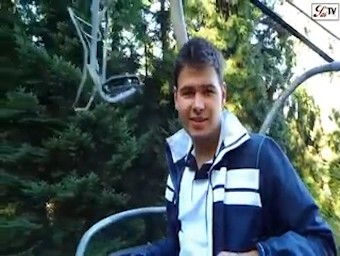}~
                \adjincludegraphics[trim={0 1.5cm 0 0.5cm},clip,valign=c,width=0.48\linewidth]{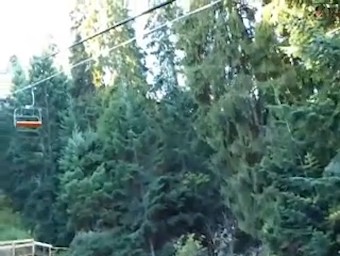}
                &\parbox[c]{1\linewidth}{\textbf{Transcript}: 
                
                You can take one like this.} 
                
            \end{tabular}
            & 
            \bgroup
            
            \begin{tabular}{@{\hskip 0mm}p{8.2cm}@{\hskip 0.8mm}}
                \raggedright \textbf{Generated captions}  \tabularnewline[1.2mm]
                \raggedright ~\colorbox[RGB]{220,230,242}{GT:} a person is riding a ski lift and speaking to us \tabularnewline[1mm]
                \raggedright ~\colorbox[RGB]{242,220,219}{No-PT:} a man is driving a motorcycle \tabularnewline[1mm]
                \raggedright ~\colorbox[RGB]{235,241,222}{MV-GPT:} a man is walking in the woods 
            \tabularnewline
            \end{tabular}
            \egroup \\ 
        \end{tabular}
    }
    \caption{Qualitative example on YouCook2 (first row) and MSR-VTT (last two rows) including a failure case (last row). \colorbox[RGB]{220,230,242}{GT:} Ground-truth caption. \colorbox[RGB]{242,220,219}{No-PT:} No multimodal pretraining.  
    \colorbox[RGB]{235,241,222}{MV-GPT:} Our model pretrained on HowTo100M.
    }
    \label{fig:youcook_qex}
\end{figure*}

\noindent\textbf{Comparisons to the State of the Art:} 
Finally, we compare MV-GPT to existing methods on all four datasets.
Table~\ref{tab:sota-youcook} compares our method to the state of the art on YouCook2, where we outperform all prior work including works pretrained on HowTo100M. 
On ViTT (Table~\ref{tab:sota-vitt}), the gap is even larger, with our model advancing the state-of-the-art by 15\% (absolute) compared to M-MASS in B-1 and M scores. 

Despite the domain gap between instructional videos in HowTo100M and general online videos in MSR-VTT, our model outperforms all existing work as shown in Table~\ref{tab:sota-msr}.
Although UniVL also pretrains both the encoder and the decoder on HowTo100M, our method achieves relative improvements of over 31\% thanks to our end-to-end training. 
Similarly, Table~\ref{tab:sota-activitynet} shows that our pretraining method achieves state-of-the-art performance on ActivityNet-Captions despite the significant domain gap.

\noindent\textbf{Qualitative Results:} 
We show  examples from YouCook2 and MSR-VTT in Figure~\ref{fig:youcook_qex}.
The first example illustrates that our model can use the visual modality to infer the term `sauce' despite the ASR error `\textit{source}' and further recognizes its name `\textit{sriracha}'.
Similarly, the second example illustrates that our approach manages to take into account both modalities jointly. 
Finally, we show a failure case in the last row in which our model fails to capture the concept `ski lift'.
A possible explanation is that the concept of a ski lift may be rarely seen in the pretraining dataset, a problem which may be alleviated by collecting more diverse pretraining videos, or incorporating external object knowledge through the use of pre-trained object detectors.

\subsection{Non-generative Video Understanding Tasks}
Although MV-GPT is a generative model and is particularly designed for multimodal video captioning, we also find that our pretraining technique learns a powerful multimodal video encoder that can be transferred easily to multiple video understanding tasks.
In particular, we show results on VideoQA, video retrieval and action classification.
For details on each task please refer to the appendix.

\noindent\textbf{VideoQA:} 
We use MV-GPT as an encoder (no \texttt{BOS} token is fed to the decoder so it only contextualizes the input tokens; see appendix for details) and the average pooled input embedding is fed to a two-layered MLP classifier to predict the answer.
The question is simply concatenated to the ASR inputs.
Following the standard protocols in \cite{yang2020just,seo2021look}, we measure the answer prediction accuracy on MSRVTT-QA~\cite{xu2017video} and ActivityNet-QA~\cite{yu2019activitynet}. 

Table~\ref{tab:vqa} compares the accuracy  of MV-GPT to existing methods that are pretrained on HowTo100M~\cite{miech2019howto100m}.
Even though MV-GPT is not designed for this particular task, our model slightly outperforms the previous state-of-the-art VQA-T~\cite{yang2020just} (which is specifically designed for VideoQA) on both datasets.

\noindent\textbf{Video Retrieval:} 
The common practice for retrieval is to train a video-text joint embedding using \textit{discriminative} losses only, typically in the form of a standard NCE loss~\cite{gutmann2010noise}, where each video clip has a single corresponding textual caption. Here we investigate whether our generative pretraining loss can provide a boost to performance.
Since each example forms two inputs-target triplets in our bidirectional framework, we apply NCE losses on both (Bi-NCE). We then add our generative pretraining loss to this framework and report results in Table~\ref{tab:retrieval_msr}. We evaluate our model with and without ASR to compare fairly to existing works.
We report recall at $k=\{1, 5, 10\}$ (R@$k$) and median rank (MdR) on MSR-VTT~\cite{xu2016msr} following the standard 9K retrieval splits~\cite{yu2018joint}.

Our first observation is that our Bi-NCE serves as a strong baseline pretraining method for retrieval. We show that adding our generative losses further improves performance by a relative 6.3\% in R@1, yielding state-of-the-art performance.
Finally, adding ASR to our multimodal encoder further improves performance by a significant margin (+ 4\%).

\noindent\textbf{Action Classification:} 
We test the visual encoder of \model on action classification following~\cite{arnab2021vivit}.
We evaluate models using top-1 classification accuracy on Kinetics 400 and 600~\cite{kay2017kinetics}.
Note that we adopt the ViViT-Base architecture with factorized encoder following \cite{arnab2021vivit}, however we use a tubelet size of $16\times16\times4$ instead of $16\times16\times2$ to reduce complexity.
We compare our model with two different initializations for the visual encoder: random and pretrained weights on ImageNet21k.
The baseline models are finetuned on the evaluation benchmarks immediately from these initializations whereas we first post-pretrain models in our MV-GPT framework and finetune for action classification.

Table~\ref{tab:kinetics} demonstrates that MV-GPT is an effective pretraining strategy for the visual encoder.
High-capacity transformer models like ViViT are challenging to train from scratch, and overfit easily as shown in the first row.
However, ViViT initialized from an MV-GPT visual encoder trained from scratch performs substantially better, obtaining absolute improvements of 24\% on Kinetics-400 (a standard video classification benchmark).
This number is close to  the performance of ViViT initalized with ImageNet-21K pretraining, as done by the original authors~\cite{arnab2021vivit} (note that ImageNet-21K was created with high manual annotation cost, while we used no labels at all during pretraining).
Finally, initialising the MV-GPT visual encoder with these same ImageNet-21K weights, and then pretraining the MV-GPT visual encoder weights on HowTo100M achieves the best results, improving upon the initialisation of~\cite{arnab2021vivit} by 1.5\% and 1.8\% on Kinetics-400 and Kinetics-600 respectively, which is the current state of the art on this dataset with this particular architecture.

\begin{table}[t]
    \centering
    \scalebox{0.8}{
    \begin{tabular}{@{\hskip 0.8mm}lcc@{\hskip 0.8mm}}
    \toprule
        Method & MSRVTT-QA & ActivityNet-QA \\
    \midrule 
        SSML~\cite{amrani2020noise} & 35.1 & --\\
        MAR-VQA~\cite{zhuang2020multichannel} & -- & 34.6\\
        DECEMBERT~\cite{tang2021decembert} & 37.4 & -- \\
        CoMVT~\cite{seo2021look} & 39.5 & 38.8\\
        VQA-T~\cite{yang2020just} & 41.5 & 38.9 \\ \hdashline
        \bf MV-GPT (Ours) & \bf 41.7 & \bf 39.1\\
    \bottomrule
    \end{tabular}}
    \caption{Comparison to SOTA on MSRVTT-QA and ActivityNet-QA for video question answering. Our method is comparable to other works, even those designed specifically for the task of VideoQA. We compare models pretrained on HowTo100M.}
    \label{tab:vqa}
\end{table}

\begin{table}[t]
    \centering
    \scalebox{0.85}{
    \begin{tabular}{@{\hskip 0.8mm}lccccc@{\hskip 0.8mm}}
    \toprule
        Method & With ASR & R@1 & R@5 & R@10 & MdR \\
    \midrule 
        UniVL~\cite{luo2020univl} & & 21.2 & 49.6 & 63.1 & 6 \\
        MMT~\cite{gabeur2020Learning} & & 26.6 & 57.1 & 69.6 & 4\\
        AVLnet~\cite{rouditchenko2021avlnet} & & 27.1 & 55.6 & 66.6 & 4\\
        SSB~\cite{patrick2021supportset} & & 30.1 & 58.5 & 69.3 & \bf3 \\
        HiT~\cite{liu2021hit} & & 30.7 & 60.9 & 73.2 & -- \\ \hdashline
        No PT & & 3.5 & 8.0 & 12.1 & 114 \\
        Bi-NCE & & 31.6 & 59.0 & 70.2 & \bf3 \\
        \bf MV-GPT (Ours) &  & \bf 33.6 & \bf 61.2 & \bf 73.6 & \bf3 \\
        \hdashline
        No PT & \checkmark & 5.6 & 13.3 & 18.4 & 92 \\
        Bi-NCE & \checkmark & 33.7 & 61.6 & 73.0 & 3 \\
        \bf MV-GPT (Ours) & \checkmark & \bf 37.3 & \bf 65.5 & \bf 75.1 & \bf2\\
    \bottomrule
    \end{tabular}}
    \caption{Comparison to SOTA on MSR-VTT for video retrieval.
    We compare models pretrained on HowTo100M. \textbf{R@$\boldsymbol{k}$}: Recall at $k$. \textbf{MdR}: Median rank. 
    }
    \label{tab:retrieval_msr}
\end{table}

\begin{table}[t]
    \centering
    \scalebox{0.85}{
    \begin{tabular}{@{\hskip 0.8mm}lcc@{\hskip 0.8mm}}
    \toprule
        ViViT initialization & Kinetics-400 & Kinetics-600 \\ 
    \midrule 
         Scratch & 50.14 & 55.47 \\  
         MV-GPT$\dagger$ & 74.20 & 77.10 \\ 
         \hdashline
         ImageNet21k~\cite{arnab2021vivit} & 78.90 & 80.62 \\     
         ImageNet21k + MV-GPT$\dagger$ & 80.40 & 82.42 \\  
    \bottomrule
    \end{tabular}}
    \caption{Action classification results on Kinetics with different ViViT initializations. MV-GPT$\dagger$ refers to a model initalised with our MV-GPT pretraining on HowTo100M with \textit{no manually annotated} labels. We use a  factorized encoder ViViT-Base following \cite{arnab2021vivit}, but use a tubelet size of $16\times16\times4$ instead of $16\times16\times2$.}
    \label{tab:kinetics}
    \vspace{-0.3cm}
\end{table}

%% file: sections/conclusion.tex
\section{Conclusion}
We present a novel generative pretraining framework for multimodal video captioning. Our bi-directional generative objective jointly trains an encoder for multimodal inputs and a decoder to generate meaningful captions, by using utterances sampled at different times in unlabelled videos. The model is trained end-to-end both during pretraining and finetuning, and achieves state-of-the-art results on multiple video captioning benchmarks as well as  on other video understanding tasks, namely VideoQA, video retrieval and action classification.



%% file: supplementary/appendix.tex
\appendix

\part{Appendix} 

In this appendix, we first provide additional experimental results and  descriptions on the dataset configurations for our pretraining and downstreams tasks in Section~\ref{sec:add_exps} and \ref{sec:supp-dataset}.
Further implementation details for the downstream tasks are described in Section~\ref{sec:supp-impl-details}.
We then present more qualitative results in Section~\ref{sec:supp-qual}.
Finally, we discuss limitations and broader impacts of our method in Section~\ref{sec:limitation} 

\section{Additional Experiments}
\label{sec:add_exps}

\subsection{Ablations on MSR-VTT}
We perform additional ablations on MSR-VTT (mirroring Table~1 in the main manuscript) and provide results in 
Table~\ref{tab:comps-loss-supp}. We observe similar trends as on YouCook2 (Table~1), albeit with smaller gaps. We believe this is due to the larger domain gap between HowTo100M and MSR-VTT. 

\subsection{Impact of Pretraining Dataset Size} \label{sec:supp-ds_size}

\begin{figure}
    \centering
    \includegraphics[width=1\linewidth]{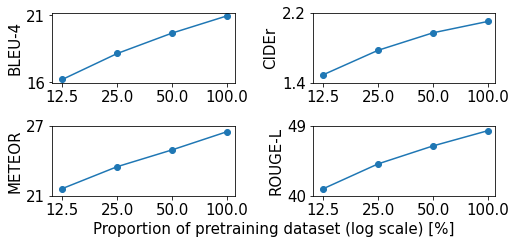}
    \caption{Performance changes in four captioning metrics with varying pretraining dataset sizes on YouCook2 for video captioning. }
    \label{fig:scale}
\end{figure}

Figure~\ref{fig:scale} reports performance against dataset size. 
All four metrics show improvement (almost linear) when the dataset size is doubled. 
This signifies that our model could improve further by collecting more unlabelled videos for pretraining.

\subsection{Open-ended Generative VideoQA}
To further investigate our model's decoding capability, we test our model on the open-ended long-form VideoQA (OL-VideoQA) benchmark \cite{zhang2019open}.
Note that the training set is smaller than the one reported in \cite{zhang2019open} (26K vs.\ 53K examples) although we obtained the dataset directly from the authors.
We test our model with/without pretraining to show its effectiveness and report the scores in B-1 and WUPS@$\alpha$ metrics where $\alpha$ is a threshold for word similarity (see \cite{zhang2019open} for details).
In Table~\ref{tab:ogvideoqa}, our model without pretraining (No PT) serves as a strong baseline outperforming almost all the scores of the existing methods despite the fewer training examples used.
The pretrained model (MV-GPT) then boosts performances further in all the metrics.
Note that the gaps in WUPS@0.0 are relatively small since all soft matches are equally weighted regardless of their semantic similarities.

\subsection{Impact of Decoder as a Part of Encoder}
\begin{figure*}
    \centering
    \includegraphics[width=\linewidth]{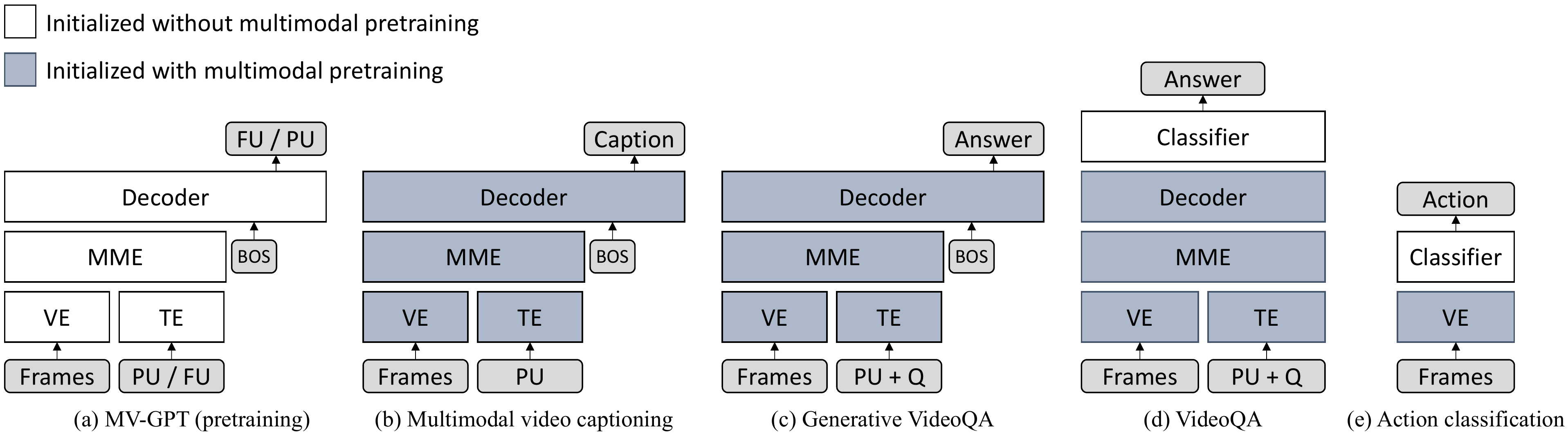}
    \caption{\textbf{Overview of pretraining (a) and finetuning (b-d)}. Each figure shows the network architecture (white and blue boxes) and the inputs and outputs (grey boxes). Blue boxes represent modules pretrained by our framework whereas white boxes are initialized without our multimodal pretraining.
    \textbf{VE:} Visual Encoder. \textbf{TE:} Text Encoder. \textbf{MME:} Multimodal Encoder. \textbf{PU:} Present Utterances. \textbf{FU:} Future Utterance. \textbf{Q:} Question. 
    }
    \label{fig:supp_transfer_arch}
\end{figure*}

As described in the main manuscript and depicted in Figure~\ref{fig:supp_transfer_arch}d, we use the pretrained decoder as a part of the encoder for the VideoQA model. 
To investigate the effectiveness of our decoder when used as a part of an encoder,
we compare our model with and without the decoder for VideoQA, and observe a 1.0\% and 0.8\% gain in accuracy with the decoder on the MSRVTT-QA and ActivityNet-QA benchmarks respectively.

\begin{table}[t]
    \centering
    \scalebox{0.85}{
     \begin{tabular}{@{\hskip 0.8mm}l@{\hskip 0.2mm}c@{\hskip 1.5mm}c@{\hskip 3mm}c@{\hskip 3mm}c@{\hskip 3mm}c@{\hskip 0.8mm}}
    \toprule
        PT Losses  & PT parts & B-4 & C & M & R-L \\
    \midrule 
 No PT & -- & 45.99	& 0.48 & 36.12 & 61.55 \\
 Baseline PT & E & 46.47 & 0.51 & 36.64 & 61.82 \\
 CoMVT & E & 47.02 & 0.52 & 37.03 & 62.19 \\
 M-MASS & E+D & 47.88 & 0.56 & 38.00 & 63.27 \\
 UniVL & E+D & 47.17 & 0.56 & 37.17 & 63.53 \\
 \bf MV-GPT & \bf E+D & \bf48.92 & \bf0.60 & \bf38.66 & \bf64.00 \\ 
    \bottomrule
    \end{tabular}}
    \caption{
    Comparisons to existing pretraining losses on MSR-VTT. 
    } 
    \label{tab:comps-loss-supp}

\end{table}

\begin{table}[t]
\vspace{-0.1cm}
    \centering
    \scalebox{0.75}{
     \begin{tabular}{@{\hskip 0.8mm}l@{\hskip 3mm}c@{\hskip 3mm}c@{\hskip 3mm}c@{\hskip 3mm}c@{\hskip 3mm}c@{\hskip 0.8mm}}
    \toprule
        Method & Train-set size & Bleu-1 & WUPS@0.9 & WUPS@0.0 \\
    \midrule 
MN+ & 19.86 & 28.37 & 56.87 \\
UNIFY & 24.13 & 29.85 & 58.56 \\
STVQA+ & 24.64 & 33.37 & 58.97 \\
CDMN+ & 52,604 & 25.38 & 34.53 & 59.20 \\
AHN & 52,604 & 25.81 & 34.14 & 59.66 \\
HCSA & 52,604 & 28.83 & 36.90 & 61.74 \\
  \midrule 
  No PT & 25,636 & 42.32 & 37.94 & 60.47\\ 
\bf MV-GPT & \bf 25,636 & \bf46.98 & \bf40.81 & \bf62.09\\ 
    \bottomrule
    \end{tabular}}
\vspace{-0.2cm}
    \caption{Comparisons to SOTA on OL-VideoQA (from \cite{zhang2019open}).
    } 
    \label{tab:ogvideoqa}
\vspace{-0.1cm}
\end{table}

\begin{figure*}[!ht]
    \centering
    \includegraphics[width=\linewidth]{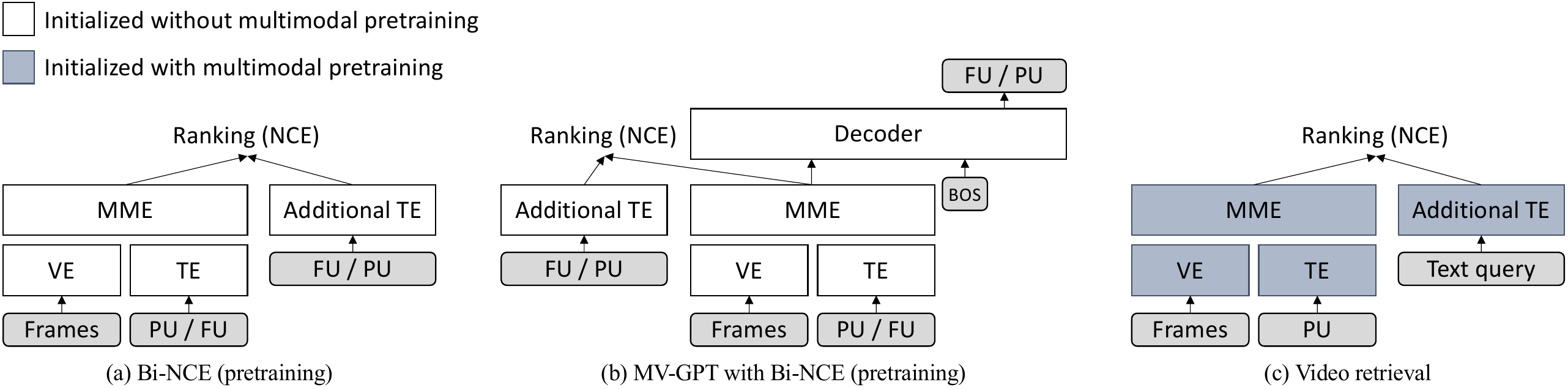}
    \caption{\textbf{Overview of pretraining with baseline Bi-NCE losses (a) and MV-GPT with Bi-NCE (b), and finetuning for video retrieval (c)}. Each figure shows the network architecture (white and blue boxes) and the inputs and outputs (grey boxes). 
    Blue boxes represent modules that are initialized with the pretrained weights whereas white boxes are trained from scratch.
    We use NCE losses to train scores for the matching pairs of a multimodal video and a text.
    Note that we use an additional text encoder to compute the target text embedding.
    \textbf{VE:} Visual Encoder. \textbf{TE:} Text Encoder. \textbf{MME:} Multimodal Encoder. \textbf{PU:} Present Utterances. \textbf{FU:} Future Utterance.}
    \label{fig:supp_retrieval}
\end{figure*}
\section{Datasets}
\label{sec:supp-dataset}
\subsection{Pretraining Dataset Preparation} 
We prepare our pretraining dataset following \cite{seo2021look} and extract triplets $(F, U, W)$ of the video frames $F$, the present utterance $U$, and the future utterance $W$, from the videos in HowTo100M~\cite{miech2019howto100m}.
We obtained transcribed speech using the YouTube ASR API~\footnote{YouTube  Data  API. \url{https://developers.google.com/youtube/v3/docs/caption}}, however these are noisy. 
To respect licensing terms, videos that have been removed from YouTube since the dataset was originally created are not used.
We then divide these videos into shorter video clips. 
The duration of video clips is determined as follows: we start with a single ASR sentence and then iteratively expand the length of the video clip backwards by adding previous sentences until the segment is longer than 5 seconds.
Each video clip therefore contains full sentences (no sentences are cut-off mid way). 
This process results in 53.5M training examples.
Since we focus on the pretraining approach, we keep only 7.5K examples as a small validation split.

\subsection{Datasets for Non-generative Tasks}
In addition to the datasets used for multimodal video captioning, which are described in the main manuscript, we make use of the following datasets for the experiments on the non-generative video understanding tasks.

\noindent\textbf{MSR-VTT}~\cite{xu2016msr} is commonly adopted for video retrieval. We follow the standard splits for retrieval \cite{yu2018joint} containing 9K and 1K examples in train and test sets, respectively.

\noindent\textbf{MSRVTT-QA}~\cite{xu2017video} is a VideoQA benchmark derived from MSR-VTT, and contains 243K QA pairs. The dataset follows the standard splits released in MSR-VTT~\cite{xu2016msr}.

\noindent\textbf{ActivityNet-QA}~\cite{yu2019activitynet} contains 58K QA pairs for VideoQA where the train, val and test sets have 32K, 18K and 8K pairs, respectively. 

\noindent\textbf{Kinetics}~\cite{kay2017kinetics} is the largest action classification benchmark.
We evaluate on both Kinetics 400 and 600, containing approximately 267K clips from 400 classes and 446K clips from 600 classes, respectively.

\section{Implementation Details} \label{sec:supp-impl-details}

\subsection{Pretraining}
As described in the main manuscript, we pretrain our model by the proposed bidirectional loss, which consists of the forward and backward generation loses.
As described in the main manuscript, our framework pretrains a model consisting of a visual encoder (VE), a text encoder (TE), a multimodal encoder (MME) and a decoder (Figure~\ref{fig:supp_transfer_arch}a).
After pretraining, a different subset of these components depending on the downstream task is transferred and finetuned, which is described in the following sections.

For pretraining, we initialize the text encoder and the decoder with the standard BERT and GPT-2 weights respectively pretrained on large-scale unlabelled corpora \cite{devlin2018bert,radford2019language}.
Similarly, we initialize our visual encoder using the pretrained weights on Kinetics 400 in \cite{arnab2021vivit}. Our entire model is pretrained end-to-end using the Adam optimizer~\cite{kingma2014adam} for 1.5M iterations with the batch size of 2048. 
We adopt a weight decaying factor of 0.01, and use the cosine learning rate decay with a linear warm-up of 500 iterations.

\subsection{Multimodal Video Captioning}
Given an MV-GPT model pretrained on HowTo100M (Figure~\ref{fig:supp_transfer_arch}a), the entire pretrained MV-GPT is transferred for multimodal video captioning as our main target task as depicted in Figure~\ref{fig:supp_transfer_arch}b.
The differences are the input and output configurations as described in the main manuscript; during pretraining, we feed present utterances (PU) as inputs predicting future utterances (FU) in forward generation and vice versa in backward generation whereas our model predicts captions given present utterances for captioning.
Note also that we feed a special \texttt{BOS} token to initiate the sentence generation from the decoder.

We finetune the entire model end-to-end for 1K iterations with an initial learning rate of 0.0001 and a batch size of 512, and use the best validation checkpoint selected based on the Meteor score. For testing, we perform beam search with a beam size of 5 as in \cite{luo2020univl}. 
Note that we initialize the decoder using the weights of GPT-2~\cite{radford2019language} when we test models trained by encoder-only pretraining methods. 

\subsection{Generative VideoQA}
Generative VideoQA requires generating an open-ended answer given multimodal video and a question. 
While a question is given as an additional text input, we simply concatenate it to the present utterance; this allows us to use the original MV-GPT model for this task without any change as depicted in Figure~\ref{fig:supp_transfer_arch}c.

\subsection{VideoQA}
Following previous work~\cite{tang2021decembert}, we formulate this task as a classification problem of predicting a predefined answer class.
Note that we simply concatenate the input question to the utterances from the clip and feed the concatenation as a single textual input.
Although we do not decode any textual outputs in this task, we still make use of the decoder as an additional multimodal encoder since our decoder is also trained to contextualize the input embeddings by applying the masked language modeling on the decoder outputs (see Section 3.1.2 in the main manuscript).
Instead of feeding the \texttt{BOS} token and predicting next tokens, we first obtain the embeddings of the inputs from the decoder, average-pool these embeddings, and feed the pooled embedding to a two-layered MLP classifier to predict the answer (Figure~\ref{fig:supp_transfer_arch}d).
Note that we use the entire pretrained model but append a randomly-initialized classifier.

For every experiment, we finetune the entire model end-to-end on the downstream benchmark for 20K iterations with a batch size of 512 and report the results using the checkpoint with the best answer accuracy.

\subsection{Action Classification}
Our goal with the experiments in action classification is to show the effectiveness of the pretrained visual encoder in MV-GPT, and therefore we simply discard all the other components and append a randomly initialized classification layer to the visual encoder as illustrated in Figure~\ref{fig:supp_transfer_arch}e. 
For finetuning, we follow all the exact evaluation protocols used in \cite{arnab2021vivit}.

\begin{figure*}
    \centering
    \scalebox{0.8}{
        \begin{tabular}{cc}
            \toprule
            \textbf{Inputs (video frames and transcript)} & \textbf{Generated captions} \\ \midrule
            
            \begin{tabular}{@{}p{12.5cm}@{}}
                \centering
                \adjincludegraphics[valign=M,width=0.325\linewidth]{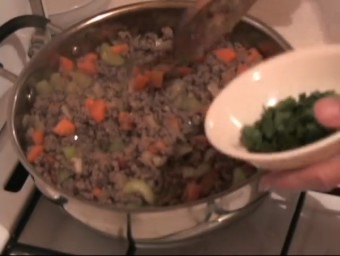}
                \adjincludegraphics[valign=M,width=0.325\linewidth]{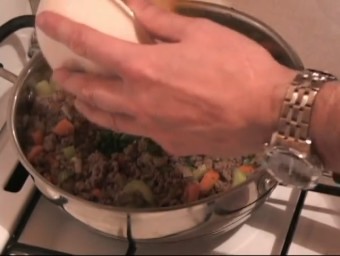}
                \adjincludegraphics[valign=M,width=0.325\linewidth]{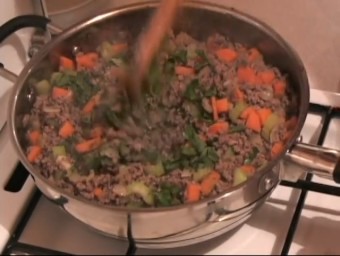} \tabularnewline[1.5cm] 
                \textbf{Transcript}: So then what I do is I add some parsley now, you don't have to put parsley do this recipe because it really doesn't call for parsing but you know something I like the atom countries and vitamins that come to the purse like unless he gives it a nice look so I just put that it but like I said, you don't have your so this about half a cup fresh firstly I use around or that's my favorite and now we just turn off the heat and we put this beside and going to be stitch.
            \end{tabular}
            & 
            \bgroup
            \def\arraystretch{1.3}
            \begin{tabular}{lp{3.8cm}}
                \raggedright \colorbox[RGB]{220,230,242}{GT:} & add parsley to the pot \tabularnewline[1.5mm]
                \raggedright \colorbox[RGB]{242,220,219}{No PT:}& add chopped tomatoes and ground beef and mix well \tabularnewline[1.5mm]
                \raggedright \colorbox[RGB]{235,241,222}{MV-GPT:}& add some chopped parsley \tabularnewline
            \end{tabular}
            \egroup \\  \midrule
            
            \begin{tabular}{@{}p{12.5cm}@{}}
                \centering
                \adjincludegraphics[valign=M,width=0.325\linewidth]{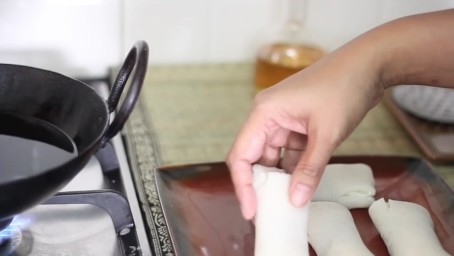}
                \adjincludegraphics[valign=M,width=0.325\linewidth]{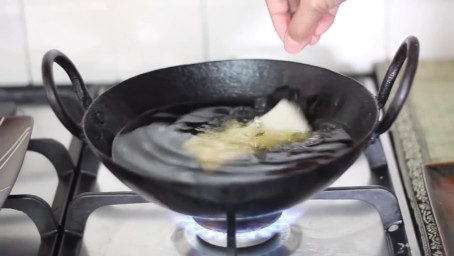}
                \adjincludegraphics[valign=M,width=0.325\linewidth]{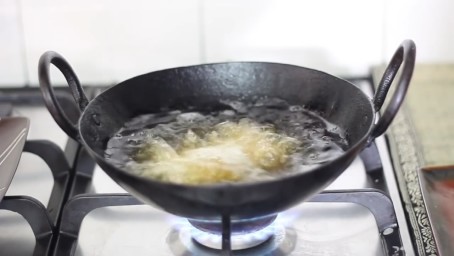} \tabularnewline[1.1cm] 
                \textbf{Transcript}: Well repeat the process and proceed to make the rest of the rolls. We will now preheat the oil for deep frying once the oil is heated add in a roll at a time and deep fry on medium heat until golden brown in color this process of deep frying takes a good 4 to 5 minutes first spring roll once browned evenly drain the excess oil and place them on a serving platter serve these delicious spring rolls with a hot and spicy Szechuan sauce these crunchy and delicious spring rolls make perfect appetizers.
            \end{tabular}
            & 
            \bgroup
            \def\arraystretch{1.3}
            \begin{tabular}{lp{3.8cm}}
                \raggedright \colorbox[RGB]{220,230,242}{GT:} & fry the rolls in oil \tabularnewline[1.5mm]
                \raggedright \colorbox[RGB]{242,220,219}{No PT:}& fry fish in oil \tabularnewline[1.5mm]
                \raggedright \colorbox[RGB]{235,241,222}{MV-GPT:}& fry the spring rolls in oil \tabularnewline
            \end{tabular}
            \egroup \\   \midrule
            
            \begin{tabular}{@{}p{12.5cm}@{}}
                \centering
                \adjincludegraphics[valign=M,width=0.325\linewidth]{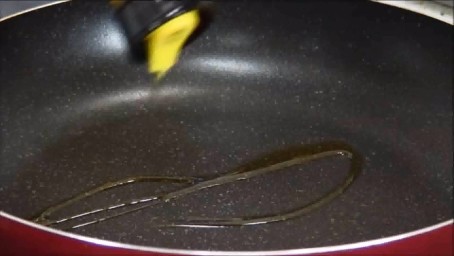}
                \adjincludegraphics[valign=M,width=0.325\linewidth]{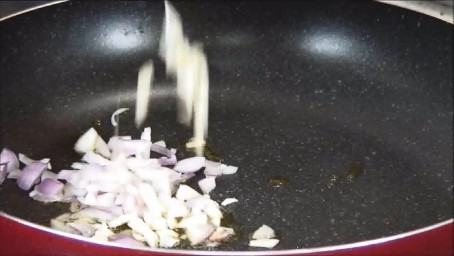}
                \adjincludegraphics[valign=M,width=0.325\linewidth]{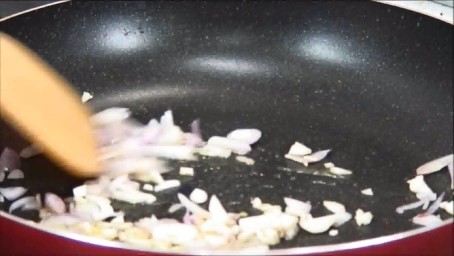} \tabularnewline[1.1cm] 
                \textbf{Transcript}: What we're going to do is add some oil to a preheated pan add the shallots in followed by the chopped garlic and I'll stir and saute this just for a minute or two until they're fragrant. Then we'll place the shrimp in followed by the fried tofu and give it a stir until the shrimp become pinkish.
            \end{tabular}
            & 
            \bgroup
            \def\arraystretch{1.3}
            \begin{tabular}{lp{3.8cm}}
                \raggedright \colorbox[RGB]{220,230,242}{GT:} & add some oil chopped shallots garlic and salt to pan \tabularnewline[1.5mm]
                \raggedright \colorbox[RGB]{242,220,219}{No PT:}& add some oil to a pan and saute \tabularnewline[1.5mm]
                \raggedright \colorbox[RGB]{235,241,222}{MV-GPT:}& add oil and shallots to a preheated pan \tabularnewline
            \end{tabular}
            \egroup \\   \midrule
            
            \begin{tabular}{@{}p{12.5cm}@{}}
                \centering
                \adjincludegraphics[valign=M,width=0.325\linewidth]{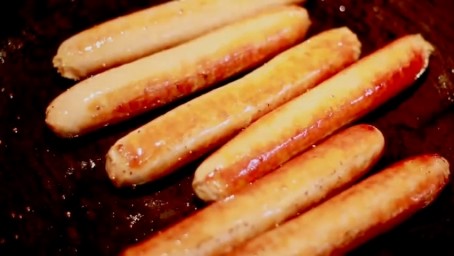}
                \adjincludegraphics[valign=M,width=0.325\linewidth]{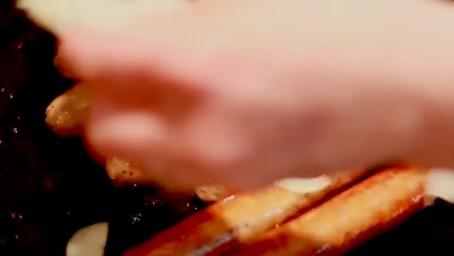}
                \adjincludegraphics[valign=M,width=0.325\linewidth]{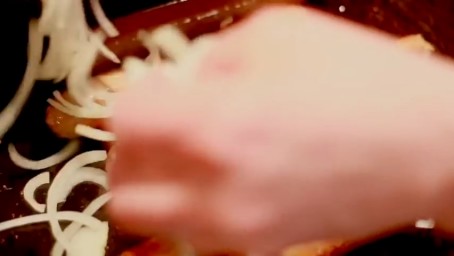} \tabularnewline[1.1cm] 
                \textbf{Transcript}: I'm just using often onion just a regular brown on him. I'm going to put that now our sausages are pretty much cook.
            \end{tabular}
            & 
            \bgroup
            \def\arraystretch{1.3}
            \begin{tabular}{lp{3.8cm}}
                \raggedright \colorbox[RGB]{220,230,242}{GT:} & add a sliced onion to the sausage pan \tabularnewline[1.5mm]
                \raggedright \colorbox[RGB]{242,220,219}{No PT:}& place the sausages on the grill \tabularnewline[1.5mm]
                \raggedright \colorbox[RGB]{235,241,222}{MV-GPT:}& add onion to the sausages \tabularnewline
            \end{tabular}
            \egroup \\   \midrule
            
            \begin{tabular}{@{}p{12.5cm}@{}}
                \centering
                \adjincludegraphics[valign=M,width=0.325\linewidth]{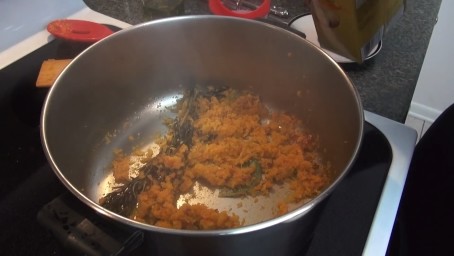}
                \adjincludegraphics[valign=M,width=0.325\linewidth]{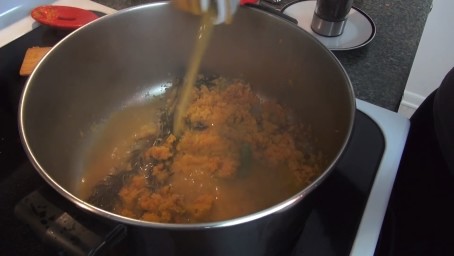}
                \adjincludegraphics[valign=M,width=0.325\linewidth]{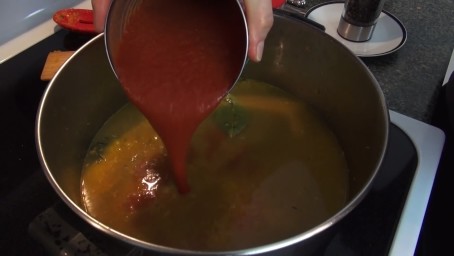} \tabularnewline[1.1cm] 
                \textbf{Transcript}: So now we're ready for the next step and I'm gonna add in the chicken broth and then you're also gonna add in your sauce at this point and the beans now just give that a star and then we're gonna raise the heat up to like medium-high and you're gonna bring this up to a boil and believe it or not.
            \end{tabular}
            & 
            \bgroup
            \def\arraystretch{1.3}
            \begin{tabular}{lp{3.8cm}}
                \raggedright \colorbox[RGB]{220,230,242}{GT:} & add in the chicken broth sauce and the beans \tabularnewline[1.5mm]
                \raggedright \colorbox[RGB]{242,220,219}{No PT:}& add crushed tomatoes to the pan and stir \tabularnewline[1.5mm]
                \raggedright \colorbox[RGB]{235,241,222}{MV-GPT:}& add chicken broth and beans to the pot \tabularnewline
            \end{tabular}
            \egroup \\   \midrule
            
            \begin{tabular}{@{}p{12.5cm}@{}}
                \centering
                \adjincludegraphics[valign=M,width=0.325\linewidth]{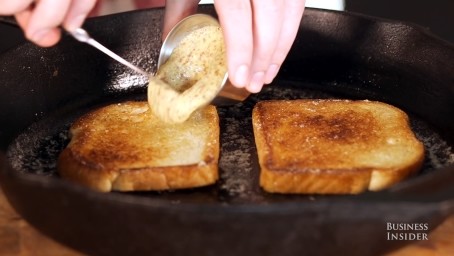}
                \adjincludegraphics[valign=M,width=0.325\linewidth]{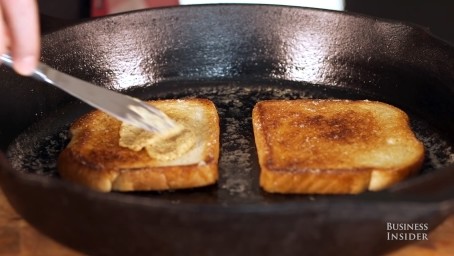}
                \adjincludegraphics[valign=M,width=0.325\linewidth]{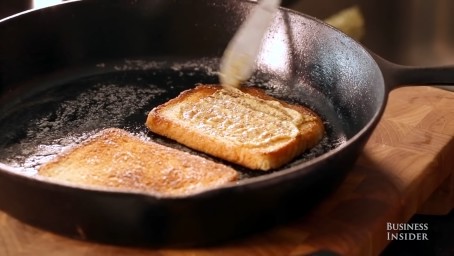} \tabularnewline[1.1cm] 
                \textbf{Transcript}: N/A
            \end{tabular}
            & 
            \bgroup
            \def\arraystretch{1.3}
            \begin{tabular}{lp{3.8cm}}
                \raggedright \colorbox[RGB]{220,230,242}{GT:} & spread mustard on the bread \tabularnewline[1.5mm]
                \raggedright \colorbox[RGB]{242,220,219}{No PT:}& flip the sandwiches \tabularnewline[1.5mm]
                \raggedright \colorbox[RGB]{235,241,222}{MV-GPT:}& spread the sauce on the bread \tabularnewline
            \end{tabular}
            \egroup \\
            
            \bottomrule
    
        \end{tabular}
    }
    
    \caption{Qualitative examples on YouCook2. \colorbox[RGB]{220,230,242}{GT.} Ground-truth caption. \colorbox[RGB]{242,220,219}{No PT.} No multimodal pretraining on HowTo100M. \colorbox[RGB]{235,241,222}{MV-GPT.} Our pretrained model. 
    }
    \label{fig:supp_youcook_ex}
\end{figure*}

\begin{figure*}
    \centering
    \scalebox{0.8}{
        \begin{tabular}{cc}
            \toprule
            \textbf{Inputs (video frames and transcript)} & \textbf{Generated captions} \\ \midrule
            
            \begin{tabular}{@{}p{12.5cm}@{}}
                \centering
                \adjincludegraphics[valign=M,width=0.325\linewidth]{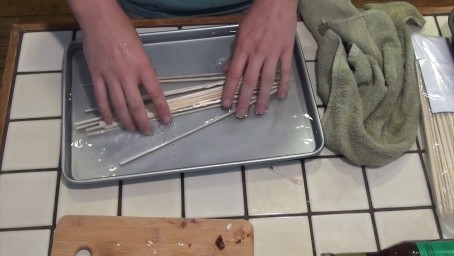}
                \adjincludegraphics[valign=M,width=0.325\linewidth]{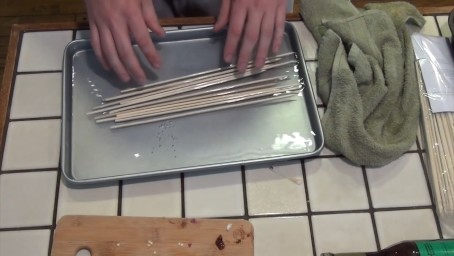}
                \adjincludegraphics[valign=M,width=0.325\linewidth]{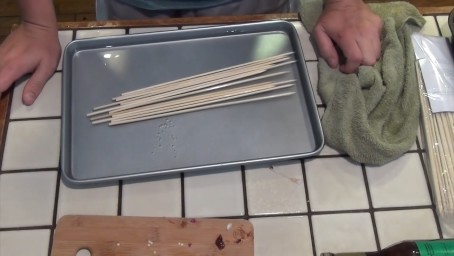} \tabularnewline[1.1cm] 
                \textbf{Transcript}: And as you can tell iris is really really excited about this at all. I'm gonna use as a shallow baking sheet. It's about half an inch thick it's about a quarter-inch full of water and the sticks will absorb a little bit of water but not much you just need to make sure you have a pan that's long enough to hold the skewers. So our skewers have been soaking for about 45 minutes the meat the vegetables and the teriyaki sauce have been also marinating for about 45 minutes.
            \end{tabular}
            & 
            \bgroup
            \def\arraystretch{1.3}
            \begin{tabular}{lp{3.8cm}}
                \raggedright \colorbox[RGB]{220,230,242}{GT:} & wash some skewers by soaking in water \tabularnewline[1.5mm]
                \raggedright \colorbox[RGB]{242,220,219}{No PT:}& soak the seaweed in water \tabularnewline[1.5mm]
                \raggedright \colorbox[RGB]{235,241,222}{MV-GPT:}& soak the skewers in water \tabularnewline
            \end{tabular}
            \egroup \\  \midrule
            
            \begin{tabular}{@{}p{12.5cm}@{}}
                \centering
                \adjincludegraphics[valign=M,width=0.325\linewidth]{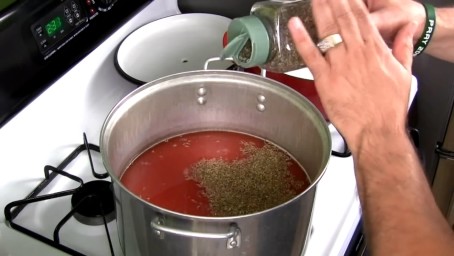}
                \adjincludegraphics[valign=M,width=0.325\linewidth]{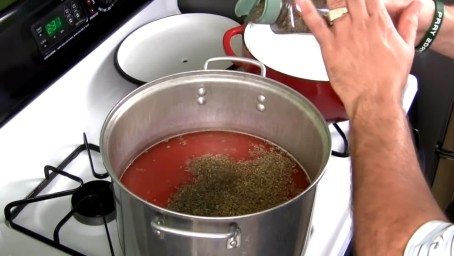}
                \adjincludegraphics[valign=M,width=0.325\linewidth]{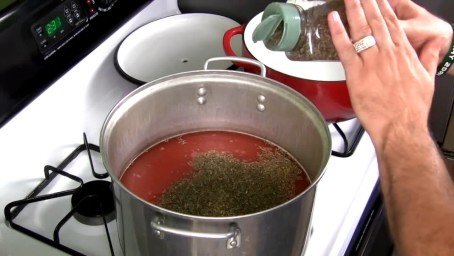} \tabularnewline[1.1cm] 
                \textbf{Transcript}: And there's the basil. There we go.
            \end{tabular}
            & 
            \bgroup
            \def\arraystretch{1.3}
            \begin{tabular}{lp{3.8cm}}
                \raggedright \colorbox[RGB]{220,230,242}{GT:} & add some basil to the pot \tabularnewline[1.5mm]
                \raggedright \colorbox[RGB]{242,220,219}{No PT:}& add paprika powder and tomato puree to the pan \tabularnewline[1.5mm]
                \raggedright \colorbox[RGB]{235,241,222}{MV-GPT:}& add basil to the pot \tabularnewline
            \end{tabular}
            \egroup \\   \midrule
            
            \begin{tabular}{@{}p{12.5cm}@{}}
                \centering
                \adjincludegraphics[valign=M,width=0.325\linewidth]{figures/supp/9_1.jpeg}
                \adjincludegraphics[valign=M,width=0.325\linewidth]{figures/supp/9_2.jpeg}
                \adjincludegraphics[valign=M,width=0.325\linewidth]{figures/supp/9_3.jpeg} \tabularnewline[1.1cm] 
                \textbf{Transcript}: So now we're ready for the next step and I'm gonna add in the chicken broth and then you're also gonna add in your sauce at this point and the beans now just give that a star and then we're gonna raise the heat up to like medium-high and you're gonna bring this up to a boil and believe it or not.
            \end{tabular}
            & 
            \bgroup
            \def\arraystretch{1.3}
            \begin{tabular}{lp{3.8cm}}
                \raggedright \colorbox[RGB]{220,230,242}{GT:} & add in the chicken broth sauce and the beans \tabularnewline[1.5mm]
                \raggedright \colorbox[RGB]{242,220,219}{No PT:}& add crushed tomatoes to the pan and stir \tabularnewline[1.5mm]
                \raggedright \colorbox[RGB]{235,241,222}{MV-GPT:}& add chicken broth and beans to the pot \tabularnewline
            \end{tabular}
            \egroup \\   \midrule
            
            \begin{tabular}{@{}p{12.5cm}@{}}
                \centering
                \adjincludegraphics[valign=M,width=0.325\linewidth]{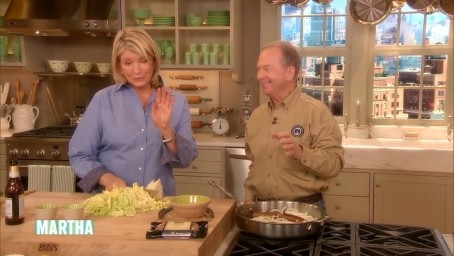}
                \adjincludegraphics[valign=M,width=0.325\linewidth]{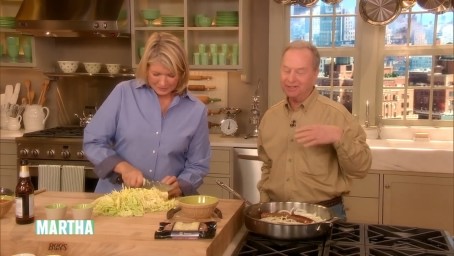}
                \adjincludegraphics[valign=M,width=0.325\linewidth]{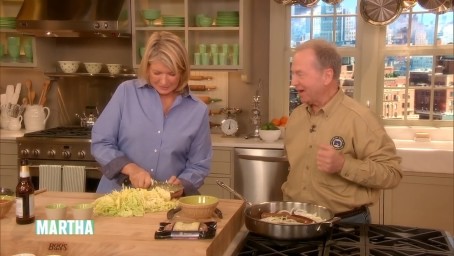} \tabularnewline[1.1cm] 
                \textbf{Transcript}: They really do taste the flavor. It's dramatically different. Really.
            \end{tabular}
            & 
            \bgroup
            \def\arraystretch{1.3}
            \begin{tabular}{lp{3.8cm}}
                \raggedright \colorbox[RGB]{220,230,242}{GT:} & continue chopping the cabbage \tabularnewline[1.5mm]
                \raggedright \colorbox[RGB]{242,220,219}{No PT:}& add salt to the pan \tabularnewline[1.5mm]
                \raggedright \colorbox[RGB]{235,241,222}{MV-GPT:}& chop the cabbage \tabularnewline
            \end{tabular}
            \egroup \\   \midrule
            
            \begin{tabular}{@{}p{12.5cm}@{}}
                \centering
                \adjincludegraphics[valign=M,width=0.325\linewidth]{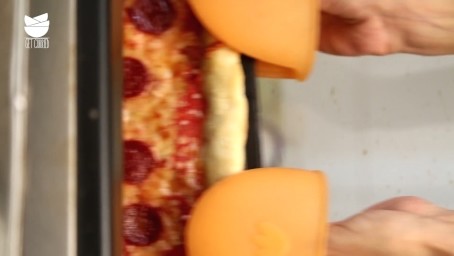}
                \adjincludegraphics[valign=M,width=0.325\linewidth]{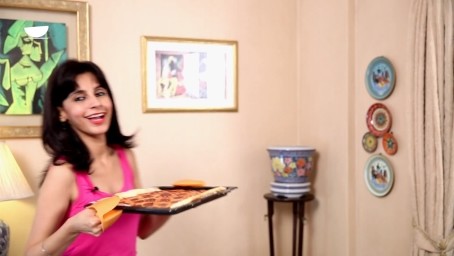}
                \adjincludegraphics[valign=M,width=0.325\linewidth]{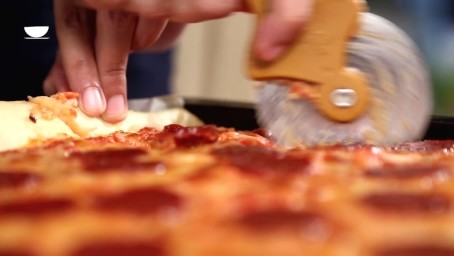} \tabularnewline[1.1cm] 
                \textbf{Transcript}: N/A
            \end{tabular}
            & 
            \bgroup
            \def\arraystretch{1.3}
            \begin{tabular}{lp{3.8cm}}
                \raggedright \colorbox[RGB]{220,230,242}{GT:} & remove from the oven and slice \tabularnewline[1.5mm]
                \raggedright \colorbox[RGB]{242,220,219}{No PT:}& bake the pizza in the oven \tabularnewline[1.5mm]
                \raggedright \colorbox[RGB]{235,241,222}{MV-GPT:}& remove the pizza from the oven and serve \tabularnewline
            \end{tabular}
            \egroup \\
            
            \bottomrule
    
        \end{tabular}
    }
    
    \caption{More qualitative examples on YouCook2. \colorbox[RGB]{220,230,242}{GT.} Ground-truth caption. \colorbox[RGB]{242,220,219}{No PT.} No multimodal pretraining on HowTo100M. \colorbox[RGB]{235,241,222}{MV-GPT.} Our pretrained model. 
    }
    \label{fig:youcook_ex2}
\end{figure*}

\subsection{Video Retrieval}

The common practice for retrieval is to train a video-text joint embedding using \textit{discriminative} losses only, typically in the form of a standard NCE loss~\cite{gutmann2010noise}, where each video clip has a single corresponding textual caption.
In the retrieval experiments, we investigate whether our generative pretraining loss can provide a boost to performance.
Since each example forms two inputs-target triplets, \ie $(F, U, W)$ and $(F, W, U)$, in our bidirectional frameworks, we apply NCE losses on both (Bi-NCE; Figure~\ref{fig:supp_retrieval}a). Note that we use an additional textual encoder to compute embeddings of the target texts. 
We then add our generative pretraining loss to this framework (Figure~\ref{fig:supp_retrieval}b).
Finally for finetuning, we transfer the visual/textual/multimodal encoders and the additional text encoder of the pretrained model, and train the network using an NCE loss with the text query provided in the downstream benchmark.

For pretraining, we down-weight the bidirectional NCE losses with a factor of 0.001, and follow the same hyper-parameters used in the regular MV-GPT pretraining.
For finetuning, we train the entire network end-to-end for 1K iterations with a batch size of 512 and we report the scores from the best checkpoints on the validation set.

\section{More Qualitative Examples} \label{sec:supp-qual}

We show more qualitative examples on YouCook2 in Figure~\ref{fig:supp_youcook_ex}. 
These qualitative examples demonstrate that our MV-GPT model can capture both textual cues (\eg, the word `parsely' in the first example) and visual cues (\eg, the action of `spreading sauce' in the last example) whereas the model without pretraining is often unable to capture these.

\section{Limitations and Broader Impact}
\label{sec:limitation}
\noindent \textbf{Limitations:} Our approach is not always successful, in particular in the presence of a significant domain shift between the pretraining data and the downstream application. Future work will address this limitation, for example by collecting curated pretraining data.

\noindent\textbf{Broader Impact:} 
Large, uncurated datasets scraped from the web may contain unintended biases, and models pretrained on such datasets may inadvertently amplify these biases.
Therefore, applications of our work beyond the academic setting presented here should first carefully examine and filter the pretraining dataset for potentially harmful biases in the data.